\documentclass[11pt]{article}
\usepackage[margin=1in]{geometry}
\usepackage{booktabs}
\usepackage{amsmath}
\usepackage{amssymb}
\usepackage{graphicx}
\usepackage{xcolor}
\usepackage[hidelinks]{hyperref}
\usepackage{microtype}
\newcommand{\pidev}{PI.DEV}
\newcommand{\citet}[1]{\cite{#1}}
\newcommand{\citep}[1]{\cite{#1}}
\newcommand{\cc}{Claude Code}
\newcommand{\xd}{$\times$}

\title{Prompt-Induced Waste in Coding Agents:\\
Reasoning, Effort, Harness Design, and End-to-End Cost\\
\large A Preregistered Benchmark with Harness Extensions}

\author{Sarel Weinberger\\
PointFive\\
\small\texttt{sarel.weinberger@pointfive.co}
\and
Amir Hozez\\
PointFive\\
\small\texttt{amir@pointfive.co}}

\date{August 2026}

\begin{document}
\maketitle

\begin{abstract}
Coding-agent efficiency cannot be characterized by token count or model price
alone. End-to-end cost and task success depend jointly on prompt semantics,
inference effort, harness policy, model, task difficulty, tool use, context
management, and provider accounting. Controlled experiments show that prompt
wording can change reasoning and verification behavior without changing the
task, that additional inference effort can help on difficult tasks but can also
add cost without benefit, and that the value of an efficiency intervention can
change when the harness changes. Prompt, effort, and harness therefore act as
interacting experimental factors rather than independent controls.
We model efficiency as cost per successful task induced by the agent trajectory.
Token and cache counts are measurements of that trajectory, not sufficient
optimization targets. Agent evaluations should therefore measure success and
end-to-end cost while controlling the system variables that determine how the
trajectory is produced.
\end{abstract}
\section{Introduction}
Coding-agent cost is usually analyzed through quantities such as input tokens,
output tokens, context size, cache use, or model price. These quantities are
important for billing, but they do not determine how much work an agent performs.
The same task can produce different numbers of reasoning steps, model turns,
tool calls, repository reads, edits, tests, and retries. We therefore ask what determines the trajectory that generates those tokens and
tool actions, rather than treating token price alone as the object of study.

We examine three controllable parts of that trajectory. The first is
\emph{prompt semantics}. Instructions can change how the agent searches,
branches, verifies, and decides to stop even when the task, model, harness, and
nominal effort setting are fixed. The second is \emph{inference effort}. More
reasoning can be useful on difficult tasks, but additional effort can also add
computation without improving the solution. The third is the \emph{agent
harness}. Harnesses determine the system prompt and tool schemas, expose or set
reasoning controls, manage context, define tool and subagent behavior, and shape
when the model is called again. These choices can change the effect of both the
prompt and the effort setting.

The prompt experiments isolate the first factor. They show that instructions
which request alternative solutions, deeper analysis, or unusually strong
certainty can increase reasoning or verification without a corresponding gain
in task success. The mechanism is behavioral rather than lexical: a longer
prompt that merely restates the task does not reproduce these effects. The
relevant variable is the work requested by the instruction.

The effort experiments test the opposite boundary. Reducing computation is not
always beneficial. On sufficiently difficult repository-scale tasks, additional
inference effort can improve success for some model--task combinations, while in
other cases it raises cost without improving outcomes. Efficiency therefore
cannot be defined as minimum reasoning or minimum tool use. The relevant
quantity is whether additional work improves the probability of solving the task
enough to justify its cost.

The harness experiments add a system-level factor. Different production agent
runtimes expose different fixed context, tool surfaces, turn structures,
verification policies, context-management rules, and reasoning defaults. Holding
the task, model, prompt, and controller logic fixed while changing the harness
can materially change both baseline computation and the effect of an effort or
prompt intervention. The value of an efficiency control is therefore conditional
on the runtime policy in which it operates.

We represent these experiments with a common model. A task, model, prompt,
effort setting, harness, and execution environment jointly induce an agent
trajectory. That trajectory determines model requests, tool actions, context
updates, tokens, cache traffic, latency, and eventually task success. Provider
prices then map the measured trajectory to billed cost. The terminal efficiency
metric is cost per successful task, not token count by itself.

The empirical program separates these factors rather than pooling them. One set
of controlled experiments isolates prompt effects under matched
model--task--harness--effort conditions. A separate hard-task campaign measures
the cost--success effect of explicit effort where baseline failures are common
enough to expose a trade-off. Cross-harness extensions then hold the model,
tasks, prompts, and intervention logic fixed while changing the runtime. These
experiments test which factors change agent work, when additional work is useful,
and which interactions must be controlled in efficiency benchmarks. The
specific benchmark suites, harness implementations, models, experimental counts,
effect sizes, confidence intervals, and implementation checks are reported in
the Experimental Setup, Results, and Appendices rather than used to define the
problem in the introduction.

The same system model also yields a concrete training problem for an adaptive
controller: condition on the observed agent state and select among a small set
of safe interventions to minimize expected remaining cost subject to a success
non-inferiority constraint (Section~\ref{sec:controller-training}).

\section{Experimental setup}
Six open-weight reasoning models ($\geq$500B-class, served by one provider)
and Claude Sonnet~5 were evaluated on 24 deterministic coding tasks under
two production agent harnesses: \pidev{} directly and \cc{} through a
pinned protocol-translation gateway, with an additional native \cc{} arm
for the closed model. Each task has a frozen fixture, visible tests, and
hidden deterministic tests that are unavailable to the agent. Eighteen
frozen prompt variants manipulate only the user instruction. Nine primary
variants preserve the objective, acceptance criteria, and test command
verbatim; Table~\ref{tab:prompts} reports the exact manipulated clauses and
Appendix~\ref{app:prompts} gives the complete templates.

Hypotheses, metrics, and decision thresholds were preregistered. The main
screening findings were evaluated on a frozen 8-task holdout, and all
comparisons are paired within task--model--harness blocks with
task-clustered bootstrap intervals. The \textbf{4{,}644 valid runs} and
\textbf{2{,}801 condition-blind trace annotations} reported for this core
benchmark refer to this controlled prompt/harness program and its associated
replications; they do \emph{not} include the separate hard-task effort
campaign in Appendix~\ref{app:swe-effort}. That SWE-bench campaign contains
150 paired model--task--effort cells and manipulates effort level rather than
prompt wording. It is used to test the boundary condition on ``less compute is
better,'' not to re-estimate the 2.4--7.4\xd{} prompt effect or the 5--30\xd{}
between-harness cost gap. A further post-registration extension changes the
harness to DeepSeek Harness (dsh)~0.1.0rc7 while holding Sonnet~5, the five
activation fixtures, prompts, hidden evaluators, controller logic, and
statistics fixed. That extension contains 75 valid comparative runs
(5 tasks $\times$ 5 arms $\times$ 3 repetitions) after an explicit baseline
activation gate and is analyzed separately in Appendix~\ref{app:dsh}.
Appendices~\ref{app:design}--\ref{app:harness} provide the full core
design, trace-annotation protocol, gateway validation, and billing
reconciliation.

\section{Prompt wording changes agent work and cost}
\label{sec:primary}
Prompt wording changed both reasoning volume and end-to-end spend under
otherwise matched conditions. The instruction to \emph{develop several
approaches, compare their trade-offs, and implement the best one} increased
reasoning tokens by 2.4--7.4\xd{} across all six open models at comparable
success, with the effect reproduced on the frozen holdout. Deep-thinking
instructions increased reasoning by 1.6--2.2\xd{} across three collection
phases, and certainty-oriented wording by 1.3--1.9\xd{}. Scope-expanding
language was the only prompt family that consistently widened code diffs.
A bounded-efficiency template specifying scope, a smallest-sufficient-change
criterion, and a stopping rule had no measurable cost penalty and reduced
reasoning for one model. By contrast, verbatim restatement of the task had
approximately unit cost, indicating that prompt content rather than prompt
length drives the observed effects in this setup. Complete per-model results
appear in Appendix~\ref{app:results}.

Because forced branching produced the largest and most consistent increase
in reasoning, we separately tested whether it improved task success. Across
878 branching runs and 225 paired blocks, there was no statistically
credible success benefit. Aggregate success was 96.5\% for baseline and
95.2\% for branching, with discordant blocks favoring baseline 26:13. On
easy cells, branching reduced success from 98.8\% to 96.4\% and increased
scope violations by approximately 3\xd{}. An exploratory $+4.8$ percentage
point effect on hard task--model cells did not survive cross-fitted
validation: the difficulty label failed to reproduce on held-out runs and
the branching effect fell to $+0.4$ percentage points. Hidden-test-only
failures were not reduced by branching (2 baseline vs.\ 3 branching), and
no annotated trace showed a discarded branch identifying a solution that a
baseline run had searched for and missed. The benchmark contains few
architecture-dependent tasks, so potential benefits of branching in that
regime remain outside the scope of the present evidence
(Appendix~\ref{app:ma-audit}).

\section{Composition of additional reasoning}
\label{sec:semantic}
For the seven models with provider-exposed reasoning text, we annotated
2{,}801 traces using a frozen rubric and a condition-blind judge. The judge
was not shown the prompt variant and was required to provide an evidence
quote for each nonzero mechanism count. Appendix~\ref{app:design} describes
the annotation and validation protocol, and Appendix~\ref{app:examples}
provides verbatim examples.

The annotations distinguish several mechanisms. Multiple-approach prompts
add approximately 3.5 candidate strategies per run, of which approximately
3 are elaborated and then discarded; the number of implemented alternatives
increases by exactly one across all models, corresponding to the selected
approach. Deep-thinking prompts add no median increase in the observable
functional units defined by the rubric: hypotheses, evidence collection,
verification, or grounding, despite 2.2\xd{} more recorded reasoning text.
Certainty-oriented prompts add redundant re-verification of already settled
facts on all six screening models. Misleading architectural hints add
hypotheses that remain unsupported by inspected code and increase
pre-edit deliberation by 4.2\xd{}; unsupported assumptions are also the only
annotated semantic marker negatively associated with success
($\rho=-0.19$).

Bounded-efficiency wording does not reduce diagnosis or final validation on
well-specified tasks. A follow-up campaign identifies a boundary condition:
on an ambiguous specification, its stop-early framing reduced evidence
seeking and produced 3/5 hidden-test failures versus 0/5 at baseline because
the relevant in-repository requirements file was not consulted
(Appendix~\ref{app:adaptive}). Accordingly, efficiency instructions should
retain an explicit requirement to inspect repository evidence when the
specification is ambiguous. These statements concern provider-exposed
reasoning traces rather than latent computation \citep{baherwani2026};
Claude Sonnet~5 is excluded from trace-level claims because its reasoning
text is not recoverable in this dataset.

\section{Mechanisms of excess cost}
\label{sec:carriers}

Joining the reasoning annotations with deterministic tool telemetry across
all 4{,}644 runs shows that the dominant mechanisms propagate cost through
different channels.

\begin{figure}[t]\centering
\includegraphics[width=\linewidth]{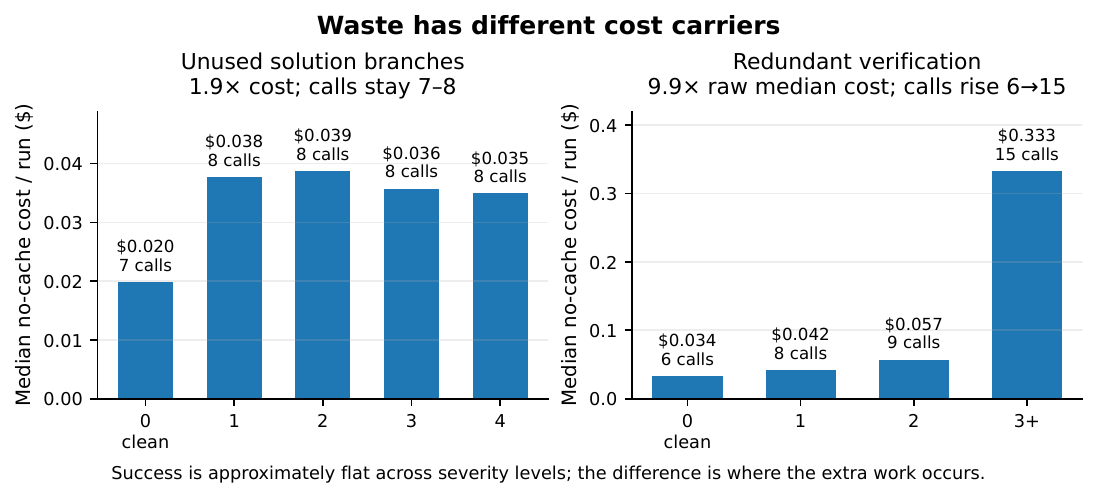}
\caption{Two waste mechanisms with different cost carriers. Bars show the
raw median no-cache cost per run; labels above each bar show the median number
of tool calls. Unused solution branches nearly double cost while calls remain
approximately flat, indicating primarily reasoning-borne overhead. Redundant
verification increases both cost and tool activity, indicating trajectory-
and tool-borne overhead. Success remains approximately flat across severity
levels.}
\label{fig:carriers}
\end{figure}

Discarded branches are primarily \textbf{token-borne}: runs with one
unused branch have $\approx$1.9\xd{} the median cost of clean runs
($n=1{,}934$ clean vs.\ 354 one-branch runs), while median tool calls
change only from 7 to 8. Code-edit counts are unchanged on all six screening
models, indicating that the additional branch exploration occurs mainly in
recorded reasoning rather than repository interaction. Redundant
verification is \textbf{tool-borne and increases with severity}:

\begin{table}[ht]\centering\small
\begin{tabular}{lrrrrrr}
\toprule
Redundant verification & $n$ & median (rel.) & mean (rel.) & wins.\ mean (rel.) & 95\% CI (median, rel.) & calls \\
\midrule
level 0 & 1{,}585 & 1.00\xd & 1.00\xd & 1.00\xd & [0.96, 1.03] & 6 \\
level 1 & 835 & 1.48\xd & 1.77\xd & 1.70\xd & [1.44, 1.58] & 8 \\
level 2 & 168 & 2.36\xd & 4.34\xd & 4.28\xd & [2.01, 2.67] & 9 \\
level 3+ & 213 & \textbf{18.25\xd} & 17.74\xd & 18.38\xd & [15.13, 21.06] & \textbf{15} \\
\bottomrule
\end{tabular}
\caption{Cost by judge-annotated redundant-verification level
($n=1{,}585$ clean vs.\ 213 high-redundancy runs). Runs at level 3+ cost
\textbf{18\xd} the clean-run median, execute 2.5\xd{} the tool calls, and
take 3\xd{} the wall-clock, with no success gradient. The gradient
survives winsorization at the 95th percentile (18.4\xd{} on winsorized
means); the level-3+ lower quartile is 10\xd{} the level-0 median.
All cost columns are normalized to the clean-run (level-0) median;
absolute values are retained in the public artifact
(\texttt{rv\_cost\_robustness.json}).}
\end{table}

Certainty-oriented wording produces this profile: $+1.75$ post-success
calls (sign-consistent on 6/6 models), extra test executions, $+4$s
latency; the most extreme observed loop re-ran an already-green suite six times.
Even local tools with no direct monetary charge create model cost: their results re-enter context
and account for a bounded 4--12\% of run cost. The synthesis:

\begin{table}[ht]\centering\scriptsize
\begin{tabular}{p{2.4cm}p{3.2cm}p{2.8cm}p{2.0cm}p{1.6cm}p{2.2cm}}
\toprule
Condition & Observable reasoning effect & Tool-layer effect & Cost carrier & Success & Evidence \\
\midrule
multiple\_approaches & $+3$ discarded branches; $+1$ implemented & weak/secondary ($p\approx.06$); edits $+0$ & token-borne & none & strong semantic; directional tool \\
deep\_thinking & 2.2\xd{} text, zero new functional units & none ($p=.26$) & token-borne & none & strong \\
max\_certainty & $+1$ redundant re-verification (6/6) & $+1.75$ post-success calls; repeat tests & tool-borne + induced tokens & none & strong \\
misleading\_hints & $+1$ unsupported assumption; 4.2\xd{} pre-edit deliberation & directional ($+1.75$ calls, $+0.5$ failed) & reasoning-borne & none / possible harm ($\rho=-0.19$) & mixed \\
bounded\_efficiency & no loss of diagnosis/validation & null ($p=1.0$) & n/a (avoids waste) & preserved & strong null \\
\bottomrule
\end{tabular}
\caption{Integrated mechanism table. ``None'' = no success gain in paired
comparison; all counts are medians of per-model paired deltas.}
\end{table}

\section{Harness effects}
\label{sec:harness}
\begin{figure}[t]\centering
\includegraphics[width=\linewidth]{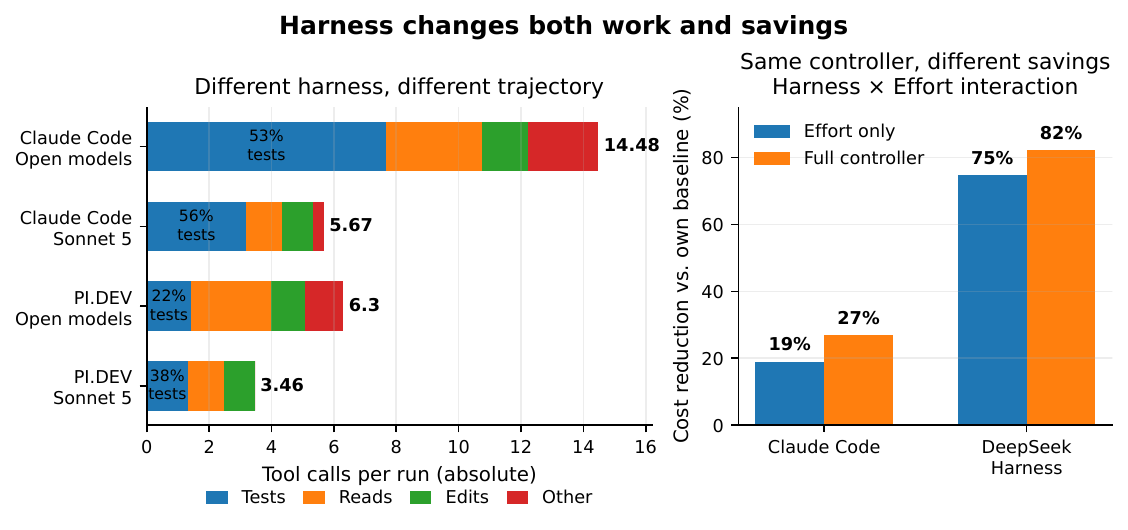}
\caption{Harness design changes both the agent trajectory and the value of
control. Left: baseline tool calls are shown in absolute calls per run rather
than normalized shares; Claude Code executes more calls and allocates a larger
fraction to tests, while PI.DEV is more inspection-heavy. ``Other'' combines
search, navigation, and residual tool categories. Right: on the same Sonnet~5
activation tasks, the same effort-control logic produces a much larger cost
reduction under DeepSeek Harness than under Claude Code. The comparison is
within each harness relative to its own baseline and illustrates the measured
harness--effort interaction.}
\label{fig:harness}
\end{figure}

Prompt effects are measured within a harness, but the between-harness
differences are larger. On matched model--task--prompt triples, \cc{}
transmits a 12--15\xd{} larger fixed prefix than \pidev{} and uses 2--7\xd{}
more turns, producing 5--30\xd{} higher cost per success at comparable
success rates. Its tool profile is verification-heavy: approximately half
of all calls are test executions, compared with 22\% under \pidev{}
(Figure~\ref{fig:harness}, left).
Reasoning traces under \cc{} also contain larger shares of planning, error
recovery, and self-correction. Prompt effects are not invariant across
harnesses: for example, goal-only prompts reduce reasoning under \cc{} but
reduce verifiability under \pidev{}. Provider-side safety classifiers do
not account for the measured billing gap: most of the comparison uses
Together-hosted open models outside Anthropic's serving path, and native
Anthropic billing reconciles from visible usage fields
(Appendix~\ref{app:harness}).

\subsection{DeepSeek Harness extension: harness changes the value of effort control}
\label{sec:dsh-main}
A post-registration extension uses DeepSeek Harness (dsh)~0.1.0rc7 with
Claude Sonnet~5 held fixed. The five tasks, prompts, fixtures, hidden judges,
controller decisions, and paired statistics are inherited from the frozen
activation benchmark; only the harness changes. Wire capture established
before comparative runs that the evaluated dsh composition enabled aggressive
reasoning by default on every ordinary request. The primary estimand is
therefore within-dsh arm versus dsh baseline, not absolute dsh versus \cc{}
cost.

\begin{table}[ht]\centering\small
\begin{tabular}{lrrr}
\toprule
Arm & Mean billed cost/run & Change vs. dsh baseline & Success \\
\midrule
baseline & \$0.2753 & n/a & 15/15 \\
policy   & \$0.2114 & $-23.2\%$ & 15/15 \\
rewrite  & \$0.2109 & $-23.4\%$ & 15/15 \\
effort   & \$0.0696 & $-74.7\%$ & 15/15 \\
full     & \$0.0490 & $-82.2\%$ & 15/15 \\
\bottomrule
\end{tabular}
\caption{DeepSeek Harness extension (Sonnet~5; $n=15$ paired runs per
arm-versus-baseline contrast). Point estimates are shown for compactness;
the preregistered paired bootstrap/McNemar/Holm procedures are retained in
the extension artifact. Component effects overlap and are not additive.}
\label{tab:dsh-main}
\end{table}

The effort-only arm accounts for approximately 91\% of the observed
baseline-to-full dollar reduction. This mechanism differs from the original
\cc{} activation benchmark, where the same controller's effort arm reduced
cost by about 19\% and the full stack by about 27\%. Under dsh the effort arm
reduced cost by about 75\% and the full stack by about 82\%. The result is a
direct empirical interaction: the benefit of effort control depends on the
harness default and request policy. The policy and rewrite arms have much
smaller point effects; the rewrite-only inferential result is not promoted as
independently conclusive. Absolute cross-harness dollar comparisons are
secondary because static prompts, tool composition, context management, and
cache behavior differ across harnesses.

The published DeepSeek Harness architecture separates the agent spine, tool
registry, persistence/checkpoint policy, compaction, subagents, workflows, and
model adapters \citep{deepseekharness}. Its tool registry can expose native
function calls or Code Mode. In Code Mode, nested tool dispatches can remain
local to execution while only the outer result enters model history
\citep{deepseekharness}. These components define additional measurable
variables: model round trips, context crossings, intermediate results added to
history, compaction events, and subagent boundaries. They were not varied in
the present extension and are not treated as experimental effects here.

\section{Model of agent cost and success}
\label{sec:system-model}
The experiments can be represented with a common trajectory model. Let
\[
\theta=(T,M,P,E,H,X,\Pi),
\]
where $T$ denotes the task (including difficulty and uncertainty), $M$ the
model, $P$ prompt semantics, $E$ explicit inference effort, $H$ the harness
configuration and policy, $X$ the execution environment and available tool
surface, and $\Pi$ the provider pricing/cache regime. The behavioral
trajectory is
\[
\tau \sim \mathcal{G}(T,M,P,E,H,X;\epsilon),
\]
a sequence of model requests, reasoning, tool actions, edits, retries, and
stopping decisions. Token and cache quantities are consequences of this
trajectory, not independent primitives. If request $k$ contains uncached
input $U_k$, cache reads $R_k$, cache writes $W_k$, and generated output
$O_k$, then a general billed-cost model is
\[
C(\tau;\Pi)=\sum_{k=1}^{N(\tau)}
\left(p_U U_k+p_R R_k+p_W W_k+p_O O_k\right)+C_{\mathrm{tool}}(\tau),
\]
where $C_{\mathrm{tool}}$ is zero for the local tools in the core benchmark
but need not be zero in a general agent system. The terminal quantity is
success-adjusted cost,
\[
\boxed{
\operatorname{CPS}(\theta)=
\frac{\mathbb{E}[C(\tau;\Pi)\mid T,M,P,E,H,X]}
{\Pr(S(\tau,T)=1\mid T,M,P,E,H,X)}
}
\]
with latency and other resource terms reported alongside it rather than
silently converted into dollars.

The dependence is not additive. A useful empirical expansion is
\[
\log \operatorname{CPS}
=\beta_0+\sum_j f_j(x_j)+\sum_{j<k}f_{jk}(x_j,x_k)+u_T+\varepsilon,
\]
where $f_{jk}$ represents pairwise interactions. The dsh extension provides
evidence for an $H\!\times\!E$ interaction; the cross-harness prompt results
show $H\!\times\!P$ variation; and the SWE-bench campaign shows variation in
$T\!\times\!E$ and $M\!\times\!E$. We do not estimate every term in this
expression. The equation specifies the variables that must be controlled or
reported when attributing an efficiency effect to a single intervention.

\section{Training an adaptive efficiency controller}
\label{sec:controller-training}
The interaction structure above makes a fixed global rule unsuitable as a
general optimization policy. A practical controller should condition on the
current agent state and select among a small set of interventions. This section
states the training problem implied by the experiments; the controller itself
is a proposed follow-up and is not evaluated in the present paper.

\paragraph{State rather than interaction tables.}
At decision time $t$, define a compact state representation
\[
s_t=\phi(T,M,H,P,E,h_t,b_t),
\]
where $h_t$ is the observed trajectory prefix and $b_t$ is the current resource
state. The representation should retain stable factors such as model and
harness, task features such as structural difficulty and specification
uncertainty, and online signals such as turns, tool calls by type, edits since
the last test, test status, repeated commands, context volume, compaction,
latency, and accrued cost. Harness identity should not be treated as a label
alone: measurable descriptors such as fixed-prefix size, tool-schema surface,
reasoning defaults, context-crossing behavior, and subagent/workflow policy are
part of the state. This lets a policy represent interactions such as
$H\!\times\!E$ or $T\!\times\!E$ without enumerating a separate rule for
every Cartesian product.

\paragraph{Small action space.}
The controller need not manipulate every measured variable. Its action set can
remain small:
\[
a_t\in\mathcal{A}=\{\text{effort},\text{prompt policy},\text{tool policy},
\text{context policy},\text{verification},\text{stop}\}.
\]
Concrete actions include lowering or escalating inference effort, removing a
branching or certainty cue, allowing or blocking a repeated tool call,
selecting which tool output re-enters context, requesting additional evidence,
and stopping after sufficient validation. Deterministic safety constraints
should define a state-dependent admissible set $\mathcal{A}_{\rm safe}(s_t)$.
For example, a green test with no intervening edit can block an identical
re-test, whereas an ambiguous specification can forbid an aggressive
stop-early action until repository evidence has been inspected.

\paragraph{Training targets.}
The training target should be outcome utility, not token reduction. Logged
executions provide tuples
\[
\mathcal{D}=\{(s_t,a_t,c_t,y,s_{t+1})\},
\]
where $c_t$ is incremental resource cost and $y$ is terminal task success. The
paired prompt, effort, and harness experiments are especially useful because
they provide matched observations of alternative policies on the same tasks.
An initial controller can fit two calibrated outcome models,
\[
\widehat c_{\rm rem}(s,a)=\mathbb{E}[C_{\rm remaining}\mid s,a],
\qquad
\widehat p_{\rm succ}(s,a)=\Pr(S=1\mid s,a),
\]
and choose
\[
\boxed{
a_t^*=\arg\min_{a\in\mathcal{A}_{\rm safe}(s_t)}
\widehat c_{\rm rem}(s_t,a)
\quad\text{s.t.}\quad
\widehat p_{\rm succ}(s_t,a)\geq p_{\min}(s_t).
}
\]
Here $p_{\min}$ is a non-inferiority floor derived from the matched baseline,
not a fixed global success threshold. This formulation prevents a controller
from appearing efficient by terminating difficult runs early.

\paragraph{Staged policy learning.}
A conservative training sequence follows directly from the evidence. First,
encode high-confidence deterministic rules for repeated successful tests,
repeated unchanged failures, and other clearly dominated actions. Second, fit
supervised cost and success models on the accumulated paired benchmark and
production telemetry. A small tabular model, gradient-boosted trees, or a
shallow network with categorical embeddings for model/harness and numeric
trajectory features is sufficient for this stage; an additional large language
model is not required for the control decision. Third, use the fitted models as
a conservative contextual policy that changes only actions with adequate data
support. Because actions alter later state, a mature version is a constrained
offline reinforcement-learning problem rather than a pure one-step classifier.
The corresponding episode-level objective is
\[
\pi^*=\arg\min_{\pi}\;\mathbb{E}_{\pi}[C(\tau;\Pi)]
\quad\text{s.t.}\quad
\Pr_{\pi}(S=1\mid z)\geq
\Pr_{\rm base}(S=1\mid z)-\delta
\]
for prespecified evaluation strata $z$ (for example model--harness and task
family), with latency or other resources added as explicit constraints when
needed. Online deployment should therefore proceed by guarded A/B evaluation
with non-inferiority checks, not by unconstrained reward maximization.

\paragraph{What the controller learns.}
The objective is not to learn that ``low effort is efficient'' or that ``fewer
tool calls are better.'' It should learn conditional policies: low effort can
be sufficient under an over-provisioned harness, higher effort can be useful on
some cross-module tasks, repeated verification after a stable green test is
usually dominated, and early stopping can be unsafe when requirements remain
uncertain. The empirical role of the state representation is to absorb these
interactions into one policy while keeping the intervention surface small and
auditable.

\section{Practical implications}
\label{sec:implications}
The experiments identify different mechanisms that can change end-to-end cost.
Token count alone does not identify which mechanism changed.

\paragraph{Prompt design.} Requests for multiple approaches are appropriate
when comparison among approaches is itself required; otherwise, the tested
formulation adds substantial deliberation without improving success.
Certainty requirements should be replaced by explicit verification and
stopping rules, for example, running the relevant suite after the final edit
and re-running it only after a relevant change or a failure. Unverified
architectural hints should be avoided. Bounded-efficiency language is
appropriate for well-specified tasks, but should explicitly require
repository inspection when the specification is incomplete or ambiguous.

\paragraph{Interpreting effect size.} Absolute per-run charges are small in
this benchmark because tasks are bounded and provider prices are low. The
main economic quantities are therefore multiplicative changes under matched
conditions. The 1.9\xd{} discarded-branch association, 18\xd{} high
redundant-verification association, 2.4--7.4\xd{} multiple-approaches effect,
and 5--30\xd{} harness gap are estimates from the controlled benchmark regimes
in which they were measured; they are \emph{not} SWE-bench effect sizes. The
separate SWE-bench campaign instead estimates the cost--success trade-off of
explicit effort levels on harder repository-scale tasks. Applying controlled-
benchmark overhead ratios to longer tasks, more expensive models, or production
workloads may imply substantial aggregate consequences, but this paper does not
make that extrapolation without an explicit workload model.

\paragraph{Harness and gateway design.} Harness configuration is the
largest cost factor measured here. Static-prefix size, tool-schema exposure,
reasoning defaults, turn count, verification policy, context compaction,
subagent/workflow topology, and which intermediate tool results re-enter model
history can all alter cost per success. Protocol translators can also
invalidate comparisons if tool schemas or reasoning items are lost; agent
benchmarks that use gateways should therefore validate multi-step tool
execution and retain provider-side usage telemetry.

\paragraph{What an efficiency benchmark must measure.} A token-only report is
not sufficient to identify the mechanism of a saving. At minimum, paired
agent evaluations should retain task success and cost per success; task
family/difficulty; model and provider; prompt variant; explicit effort;
harness/version/configuration; system-prefix and tool-schema surface; model
round trips and turns; tool calls by type; raw and delivered tool volume;
reasoning/output tokens where exposed; uncached input, cache writes, and cache
reads; context-compaction events; retries, timeouts, and empty patches;
subagent/workflow boundaries; wall-clock latency; and evidence that the
intended intervention actually activated. These fields distinguish a true
reduction in redundant work from early failure, cache-price effects, shifted
work into another channel, or a harness default that simply over-provisions
compute.

\section{Related Work}

\emph{Prompt-induced agent behavior}: the present study treats user-prompt
wording as an experimental factor and isolates its effect under matched
coding-agent conditions using a paired design over two production harnesses. \emph{Reasoning length and test-time compute}: work on
overthinking studies how much models deliberate; we measure what
deliberation is made of and what it triggers. \emph{Faithfulness of
chain-of-thought}: \citet{baherwani2026} show visible traces
under-represent computation; we adopt their implication as an epistemic
scope on all trace claims. \emph{Tool selection and efficiency}:
When2Tool \citep{sun2026when2tool} shows models latently know when a call
is necessary and suppresses unnecessary calls. \emph{Cost-aware
planning}: CostBench \citep{liu2025costbench} benchmarks cost-optimal
plan adaptation; CATP-LLM \citep{wu2024catp} trains cost-aware tool
planning; \citet{yang2026efficient} survey efficiency levers across
memory, tool learning, and planning. \emph{Context optimization in coding
agents}: CORVUS \citep{zheng2026corvus} reduces the context cost of
file-read observations. These lines of work primarily address when tools should be called, how tool
plans should trade off cost and utility, or how observations should be
compressed. DeepSeek Harness \citep{deepseekharness} is complementary
systems work: it makes orchestration choices explicit through composable
capabilities for tool presentation, compaction, subagents, workflows,
persistence, and model adapters. Its Code Mode is especially relevant to the
present cost model because multiple tool sub-dispatches can be executed behind
one model-facing result, changing context crossings without necessarily
changing the underlying tool work. The present study adds controlled evidence
that prompt, effort, and harness policy interact in their effect on observable
reasoning, downstream tool execution, success, and end-to-end cost.

\section{Limitations}

Visible reasoning is not necessarily faithful or complete
\citep{baherwani2026}; semantic labels describe observable text under a
frozen rubric, never hidden cognition. Claude Sonnet~5 has no recoverable
reasoning trace in this dataset (empty thinking blocks; reasoning billed
inside output) and is excluded from all trace-level claims; its inclusion
is limited to cost, success, turn, latency, and deterministic tool-level
analyses. Reasoning-text availability is provider- and gateway-dependent.
Exact per-turn cost attribution is unavailable; tool-induced model cost
is a bounded estimate (4--12\%). No direct tool API charge exists in this
benchmark; local CPU/GPU/memory telemetry was not recorded; \pidev{}
lacks per-call timing. ``Post-success'' rests on the visible-test
completion proxy, not hidden-evaluator satisfaction. Tool counts are
heavy-tailed and zero-inflated (94\% of runs have zero post-green repeat
tests); H-T1 and H-T4 tool-layer effects are reported as directional rather
than confirmatory. Judge validation was useful but imperfect:
researcher-blind labels agreed within $\pm1$ on 84.9\% of fields;
same-model prompt sensitivity $\kappa=0.44$; cross-judge-model agreement
was lower overall ($\kappa=0.28$) though 0.55--0.68 on
hypothesis-critical fields. Tasks are small ($\leq$4 files) with high
success ceilings; the token-borne character of branch waste may weaken on
repository-scale tasks where alternatives demand real exploration, and
findings may not transfer unchanged to long-horizon production agents.
Harness and model effects may interact; the \cc{} arms are smaller by
design. Cost figures use pinned catalog prices at collection time. Finally, whether forced branching helps on
architecture-dependent tasks is untested: the task set contains
essentially no such tasks, success ceilings limit detectable effects to
$\gtrsim$2.8pp, and the one suggestive subgroup signal failed
cross-fitted validation (Appendix~\ref{app:ma-audit}); a follow-up
adaptive-compute campaign could not create baseline failures even with
purpose-built deceptive tasks, leaving evidence-triggered escalation untestable
at this task scale (Appendix~\ref{app:adaptive}).
Provider-side safety classifiers may affect \emph{latency}: their
server-side execution time is not exposed in usage fields, so wall-clock
time cannot be decomposed into classification, queueing, and generation
components, and the absence of unexplained billing is not evidence that
no classifier executes.

The dsh extension has additional scope limits. It uses one closed model
(Sonnet~5) and five activation fixtures whose comparative arms all solve the
task; it therefore demonstrates over-provisioned effort in that harness
configuration, not the optimal effort policy on hard tasks. It does not test
DeepSeek models. The evaluated unattended SDK composition also has no
plug-and-play external pre-tool hook arm equivalent to the \cc{} runtime hook;
a custom harness plugin would be a different intervention. Absolute dsh versus
\cc{} cost is not a primary estimand because their prompts, tools, compaction,
and cache traffic differ. Finally, dsh capabilities such as Code Mode,
workflow orchestration, alternative subagent backends, and compaction policy
were not factorially varied. The highest-value follow-up is therefore to run
the already-frozen hard SWE-bench slice under dsh with both Sonnet~5 and
Opus~5, crossing harness and effort on identical tasks, followed by targeted
ablations of context-crossing and orchestration policies. The adaptive
controller in Section~\ref{sec:controller-training} is a design implication,
not an evaluated result: the present data do not establish that the proposed
outcome models, contextual policy, or offline-RL stage will improve the
cost--success frontier in deployment.

\section{Conclusion}
The experiments support a system-level account of coding-agent efficiency.
Prompt semantics can change reasoning and verification under otherwise matched
conditions. Explicit inference effort can improve success on some hard tasks,
but its value depends on the task and model. Harness policy can change both the
baseline amount of computation and the effect of an effort or prompt
intervention. These factors therefore cannot be evaluated independently.

A useful unit of analysis is the agent trajectory. Task, model, prompt, effort,
harness, and execution environment determine the sequence of model requests,
tool actions, context updates, retries, and stopping decisions. Token volume,
cache traffic, latency, and billed cost follow from that sequence. An
intervention that reduces one token category can still increase end-to-end cost
if it causes more turns, more tool calls, or additional reasoning; an
intervention that reduces effort can lower success if the task requires more
evidence or search.

The appropriate terminal metric is therefore cost per successful task, with
prompt, effort, harness configuration, tool behavior, context management,
latency, and provider usage accounting retained as explanatory variables. The
same framework specifies how an adaptive controller should be trained: encode
the current agent state, predict remaining cost and success under a small set of
safe actions, and minimize expected cost subject to a success non-inferiority
constraint. The immediate experiment is to cross the frozen hard SWE-bench
slice with both Sonnet~5 and Opus~5 under DeepSeek Harness; the resulting
matched data can then support conservative training and evaluation of the
state-conditioned controller described in Section~\ref{sec:controller-training}.
Reproduction materials, including fixtures, hidden evaluators, prompts,
runners, telemetry, and analysis code, are available in the public repository.

\appendix

\section{Full Experimental Design, Preregistration, and Metrics}
\label{app:design}

\subsection{Preregistration}
Hypotheses (H1--H8, H12--H17), metric definitions, waste-classification
rules, and experiment protocols were frozen in the repository before any
benchmark result was inspected. The holdout phase additionally froze
per-model variant selections, prompts, evaluators, and analysis code
before any holdout run. The Kimi-K3 replication froze its task rule,
matrix, and material-difference thresholds in a commit that predates its
first result.

\subsection{Models, harnesses, and serving}
Six reasoning models ($\geq$500B-class) served by Together~AI were frozen
before data collection: DeepSeek-V4-Pro, Kimi-K2.6, Kimi-K2.7-Code,
Nemotron-3-Ultra-550B-A55B, Inkling, and GLM-5.2. Two harnesses operate
them: \pidev{}~0.82.1 speaking OpenAI chat completions directly, and
\cc{}~2.1.220 speaking the Anthropic Messages protocol through a pinned
LiteLLM~1.93.0 gateway that translates to chat completions. Two logging
reverse proxies capture both sides of the gateway with secrets redacted,
so every run retains the provider's raw usage object.

\subsection{Tasks}
24 deterministic coding tasks (8 low / 8 medium / 8 high complexity;
9~Python, 9~JavaScript, 6~Go), each with a clean fixture, one concrete
objective, explicit acceptance criteria, visible tests, \emph{hidden}
deterministic tests that never exist in the workspace during a run,
allowed/forbidden paths, and a per-task evaluator. 16 tasks form the
development split; 8 were frozen as holdout. Every fixture was verified to
fail its visible tests before the fix.

\subsection{Prompt variants}
A generator renders 18 variants per task from frozen templates. Nine
\emph{primary} variants preserve the objective, acceptance criteria, and
test command verbatim (validated automatically): a precise
\texttt{baseline}; \texttt{verbose\_repetition};
\texttt{deep\_thinking}; \texttt{exhaustive\_exploration};
\texttt{multiple\_approaches}; \texttt{max\_certainty};
\texttt{adjacent\_cleanup}; \texttt{no\_questions\_autonomy};
\texttt{bounded\_efficiency} (scope + smallest-sufficient-change + stop
condition); plus \texttt{goal\_only} and \texttt{scoped\_authorization}
for the harness axes. Seven \emph{stress} variants intentionally break
semantic equivalence (ambiguity, conflicting constraints, irrelevant
context, misleading hints, multi-turn splitting and restatement) and are
analyzed separately.

\subsection{Metrics}
Behavioral: reasoning tokens (provider-reported,
\texttt{completion\_tokens\_details.reasoning\_tokens}), visible output
tokens, tool calls by type, turns, files inspected, duplicate reads,
repeated searches, repeated test runs, time to first edit, wall time,
out-of-scope changes (git diff vs.\ allowed paths), and success on visible
plus hidden tests. Billing: uncached and cached input tokens, logical
input ($=$ uncached $+$ cached), actual cost at pinned catalog prices, and
estimated no-cache cost. The primary outcome is the \emph{paired
reasoning ratio}: a run's reasoning tokens divided by the median baseline
reasoning of the same model--harness--task block. Task-clustered bootstrap
yields 95\% intervals. A variant is classified \emph{wasteful} only when
the median ratio exceeds 1.5 with CI lower bound above 1.1, no material
success gain, and the effect appears on multiple tasks.

\subsection{Reasoning-trace availability and semantic annotation}
A five-tier audit of run artifacts was completed before the semantic
annotation protocol was specified. Full
provider-exposed reasoning text is recorded for all seven open-weight
models on both harnesses (\pidev{} turn events; \cc{} transcripts via the
gateway's translation of the provider \texttt{reasoning} field, verifiable
against raw wire captures). \textbf{Claude Sonnet~5 has no recoverable
reasoning text}; its thinking blocks are empty and reasoning is billed
inside output accounting. It is therefore included in cost, success, turn,
latency, and deterministic tool-level analyses, and \emph{excluded} from
all trace annotation; we neither estimate its reasoning units nor infer
its reasoning structure from tool behavior. A frozen rubric (14
composition counts, 6 waste mechanisms, ordinal quality scores; hypotheses
H-S1--H-S6) was committed before any annotation. A condition-blind judge
(never shown variant text or names) with grammar-constrained JSON output
and \emph{mandatory evidence quotes} annotated 2{,}801 runs (98.9\% valid;
91.1\% of nonzero counts evidence-backed). Validation: researcher-blind
hand labels agreed within $\pm1$ on 84.9\% of fields; same-model
prompt-sensitivity $\kappa=0.44$; cross-judge-model $\kappa=0.28$ overall
but 0.55--0.68 on the hypothesis-critical fields (unused branches,
alternatives, redundant verification). Judge-sensitive labels
(task restatements, post-solution reasoning) are carried by deterministic
proxies instead.

\subsection{Tool taxonomy and cost decomposition}
A second frozen rubric fixes a deterministic tool taxonomy (navigation,
search, read, edit, test, build, lint, git, environment, recovery),
conservative redundancy rules (duplicate normalized commands with no
intervening edit; re-reads of unchanged files; re-runs of already-green
test scopes), and a completion proxy (last edit \emph{and} first
fully-green visible test); hidden-evaluator mid-run satisfaction is not
recoverable. Unit tests fix these rules. Blind, hand-derived labels on complete traces
were then used to validate the extractor; this validation identified a TAP
pass-detection error that was corrected before condition-aware analysis. Cost is
decomposed into model cost (ledger), \emph{direct tool cost} (none exists:
every tool is local, so direct tool cost is zero; counts and wall-clock are
reported separately), and \emph{tool-induced model cost}: tool results
re-entering context, bounded below and above because exact per-turn
billing attribution is impossible. Count variables are zero-inflated and
heavy-tailed (e.g.\ 94\% of runs have zero post-green repeat tests, max
6); analyses use per-task paired medians with task-clustered bootstrap and
sign-flip permutation tests; pooled correlations are not used as standalone
evidence.

\paragraph{Methodological consequences.} Preregistration and blinded
validation changed the interpretation in three material cases: a screening
signal failed on holdout, a replication threshold was triggered by a
baseline-denominator shift that ratios alone obscured, and a hand-label
validation pass identified an extraction error before condition-aware
analysis. These checks also support treating the observed prompt effects as
mechanism-specific rather than as a single uniform overhead.

\section{Tasks and Exact Prompts}
\label{app:taskprompts}
\paragraph{Exact wording.} Because prompt wording is the manipulated
variable, condition names alone are insufficient for reproduction: lexical
form, ordering, length, and position may themselves affect model behavior.
Table~\ref{tab:prompts} therefore reproduces the exact frozen clause for
every principal condition (byte-exact, SHA-256-verified against the run
ledger by \texttt{paper/make\_prompt\_appendix.py}, with a regression test
guarding staleness); Appendix~\ref{app:prompts} contains all 18 complete
templates, the stress-family examples, the multi-turn sequences, and a full
before-and-after rendering of one task. Our findings apply to these tested
formulations; post-registration analyses test semantic versus lexical generalization
separately; the primary benchmark itself does not assume invariance across
paraphrases.

\begin{table}[p]\centering\scriptsize
\begin{tabular}{p{2.5cm}p{9.2cm}p{3.2cm}}
\toprule
Condition & Exact manipulated clause & Intended manipulation \\ \midrule
\texttt{baseline} & {\scriptsize\ttfamily \{OBJECTIVE\}\ Scope: \{SCOPE\}\ Acceptance criteria: \{ACCEPTANCE\_CRITERIA\}\ Verify with: \{TEST\_COMMAND\}\ Stop when the tests pass.} & precise control (objective, scope, criteria, test command, stop condition) \\ \addlinespace
\texttt{multiple\_approaches} & {\scriptsize\ttfamily Develop several distinct approaches to this fix, compare their trade-offs explicitly, and only then select and implement the best one.} & require explicit comparison among multiple candidate approaches \\ \addlinespace
\texttt{deep\_thinking} & {\scriptsize\ttfamily Think very deeply about this. Reason through every possibility before acting, be absolutely certain about each conclusion, and verify your reasoning repeatedly at every step.} & demand deep, repeatedly verified reasoning \\ \addlinespace
\texttt{exhaustive\_exploration} & {\scriptsize\ttfamily Before editing anything, inspect the entire repository and make sure you understand the complete system: read every source file and understand how all the parts connect, so your change is informed by the whole codebase.} & demand whole-repository understanding first \\ \addlinespace
\texttt{max\_certainty} & {\scriptsize\ttfamily You must be absolutely confident in the result. Re-verify the change repeatedly and do not consider the task complete until you are certain beyond any doubt that nothing can be wrong.} & demand certainty beyond doubt / re-verification \\ \addlinespace
\texttt{adjacent\_cleanup} & {\scriptsize\ttfamily While you are in there, feel free to clean up anything adjacent that could be improved: refactor awkward code, tidy naming, and make related general improvements where you see the opportunity.} & authorize scope expansion \\ \addlinespace
\texttt{no\_questions\_autonomy} & {\scriptsize\ttfamily Do not ask me any questions. Infer any missing details yourself and do whatever is necessary to complete the task autonomously.} & prohibit questions and require autonomous inference \\ \addlinespace
\texttt{bounded\_efficiency} & {\scriptsize\ttfamily Work efficiently: begin with the failing test and the most likely implementation files; inspect additional files only when evidence requires it; avoid unrelated cleanup; make the smallest sufficient change; run the relevant tests; stop as soon as the acceptance criteria pass.} & explicit scope + smallest change + stop rule \\ \addlinespace
\texttt{verbose\_repetition} & {\scriptsize\ttfamily To restate the requirements so nothing is missed: the change must stay within the scope given above (shop/discount.py only), it must satisfy the same acceptance criteria already listed, and the verification command remains python3 -m unittest discover -s tests -t . -v. In other words, do the task exactly as described, within exactly the same boundaries, checked by exactly the same tests, and finish under exactly the same stop condition as stated before.} & restate identical requirements (length control) \\ \addlinespace
\texttt{goal\_only} & {\scriptsize\ttfamily (baseline reduced to \{OBJECTIVE\} only; scope, criteria, test command, stop condition removed)} & retain the objective only \\ \addlinespace
\texttt{scoped\_authorization} & {\scriptsize\ttfamily You are authorized to modify only: shop/discount.py only You are NOT authorized to modify anything else; treat all other paths as read-only. \emph{[replaces:]} Scope: shop/discount.py only} & recast scope as explicit authorization \\ \addlinespace
\texttt{misleading\_architecture} & {\scriptsize\ttfamily Note: we believe the problem is caused by the test fixtures in tests/ \char45\char45\char45 start there. \emph{[replaces:]} Scope: shop/discount.py only} & provide an incorrect architectural hint (stress) \\ \addlinespace
\bottomrule\end{tabular}
\caption{Principal prompt conditions with the exact frozen wording (byte-exact; SHA-256-verified against the run ledger). Exact manipulated clause shown; complete rendered templates appear in Appendix~\ref{app:prompts}. Placeholders are instantiated per task, identically across variants of the same task.}
\label{tab:prompts}\end{table}

\section{Exact Prompt Templates and Experimental Manipulations}
\label{app:prompts}
All wording below is reproduced byte-exactly from the frozen prompt
files used in the experiments (task \texttt{py-low-01} as the
fixed representative rendering; placeholders \{OBJECTIVE\},
\{ACCEPTANCE\_CRITERIA\}, \{TEST\_COMMAND\} vary per task and are
instantiated identically across variants of the same task). SHA-256
hashes of every displayed prompt match the hashes recorded in the
run ledger. \texttt{<TURN-BREAK>} marks the protocol's fixed
multi-turn boundary.
\subsection{Primary semantically controlled variants}
\paragraph{\texttt{baseline}}
Family: semantically controlled; hypothesis: baseline; multi-turn: no; objective/criteria/test preserved verbatim: yes; SHA-256 \texttt{e48b2d5b1f81} (ledger-verified).
\emph{Complete rendered prompt:}\begin{quote}\small\ttfamily Fix the discount bug in shop/discount.py: apply\_discount(price, pct) must return the price AFTER applying a pct-percent discount, not the discount amount itself.\\ Scope: shop/discount.py only\\ Acceptance criteria: apply\_discount(100, 20) == 80.0 and every test under tests/ passes\\ Verify with: python3 -m unittest discover -s tests -t . -v\\ Stop when the tests pass.\end{quote}
\paragraph{\texttt{verbose\_repetition}}
Family: semantically controlled; hypothesis: H1; multi-turn: no; objective/criteria/test preserved verbatim: yes; SHA-256 \texttt{1f96a9bf4320} (ledger-verified).
\emph{Text added relative to baseline (placement: after the baseline task block):}\begin{quote}\small\ttfamily To restate the requirements so nothing is missed: the change must stay within the\\ scope given above (shop/discount.py only), it must satisfy the same acceptance criteria already\\ listed, and the verification command remains python3 -m unittest discover -s tests -t . -v. In other words, do the\\ task exactly as described, within exactly the same boundaries, checked by exactly\\ the same tests, and finish under exactly the same stop condition as stated before.\end{quote}
\emph{Complete rendered prompt:}\begin{quote}\small\ttfamily Fix the discount bug in shop/discount.py: apply\_discount(price, pct) must return the price AFTER applying a pct-percent discount, not the discount amount itself.\\ Scope: shop/discount.py only\\ Acceptance criteria: apply\_discount(100, 20) == 80.0 and every test under tests/ passes\\ Verify with: python3 -m unittest discover -s tests -t . -v\\ Stop when the tests pass.\\ To restate the requirements so nothing is missed: the change must stay within the\\ scope given above (shop/discount.py only), it must satisfy the same acceptance criteria already\\ listed, and the verification command remains python3 -m unittest discover -s tests -t . -v. In other words, do the\\ task exactly as described, within exactly the same boundaries, checked by exactly\\ the same tests, and finish under exactly the same stop condition as stated before.\end{quote}
\paragraph{\texttt{deep\_thinking}}
Family: semantically controlled; hypothesis: H2; multi-turn: no; objective/criteria/test preserved verbatim: yes; SHA-256 \texttt{d8be42655ded} (ledger-verified).
\emph{Text added relative to baseline (placement: after the baseline task block):}\begin{quote}\small\ttfamily Think very deeply about this. Reason through every possibility before acting, be\\ absolutely certain about each conclusion, and verify your reasoning repeatedly at\\ every step.\end{quote}
\emph{Complete rendered prompt:}\begin{quote}\small\ttfamily Fix the discount bug in shop/discount.py: apply\_discount(price, pct) must return the price AFTER applying a pct-percent discount, not the discount amount itself.\\ Scope: shop/discount.py only\\ Acceptance criteria: apply\_discount(100, 20) == 80.0 and every test under tests/ passes\\ Verify with: python3 -m unittest discover -s tests -t . -v\\ Stop when the tests pass.\\ Think very deeply about this. Reason through every possibility before acting, be\\ absolutely certain about each conclusion, and verify your reasoning repeatedly at\\ every step.\end{quote}
\paragraph{\texttt{exhaustive\_exploration}}
Family: semantically controlled; hypothesis: H2; multi-turn: no; objective/criteria/test preserved verbatim: yes; SHA-256 \texttt{7fa4f53c909e} (ledger-verified).
\emph{Text added relative to baseline (placement: after the baseline task block):}\begin{quote}\small\ttfamily Before editing anything, inspect the entire repository and make sure you understand\\ the complete system: read every source file and understand how all the parts\\ connect, so your change is informed by the whole codebase.\end{quote}
\emph{Complete rendered prompt:}\begin{quote}\small\ttfamily Fix the discount bug in shop/discount.py: apply\_discount(price, pct) must return the price AFTER applying a pct-percent discount, not the discount amount itself.\\ Scope: shop/discount.py only\\ Acceptance criteria: apply\_discount(100, 20) == 80.0 and every test under tests/ passes\\ Verify with: python3 -m unittest discover -s tests -t . -v\\ Stop when the tests pass.\\ Before editing anything, inspect the entire repository and make sure you understand\\ the complete system: read every source file and understand how all the parts\\ connect, so your change is informed by the whole codebase.\end{quote}
\paragraph{\texttt{multiple\_approaches}}
Family: semantically controlled; hypothesis: H3; multi-turn: no; objective/criteria/test preserved verbatim: yes; SHA-256 \texttt{820999071ae6} (ledger-verified).
\emph{Text added relative to baseline (placement: after the baseline task block):}\begin{quote}\small\ttfamily Develop several distinct approaches to this fix, compare their trade-offs\\ explicitly, and only then select and implement the best one.\end{quote}
\emph{Complete rendered prompt:}\begin{quote}\small\ttfamily Fix the discount bug in shop/discount.py: apply\_discount(price, pct) must return the price AFTER applying a pct-percent discount, not the discount amount itself.\\ Scope: shop/discount.py only\\ Acceptance criteria: apply\_discount(100, 20) == 80.0 and every test under tests/ passes\\ Verify with: python3 -m unittest discover -s tests -t . -v\\ Stop when the tests pass.\\ Develop several distinct approaches to this fix, compare their trade-offs\\ explicitly, and only then select and implement the best one.\end{quote}
\paragraph{\texttt{max\_certainty}}
Family: semantically controlled; hypothesis: H4; multi-turn: no; objective/criteria/test preserved verbatim: yes; SHA-256 \texttt{1b35cdaf8727} (ledger-verified).
\emph{Text added relative to baseline (placement: after the baseline task block):}\begin{quote}\small\ttfamily You must be absolutely confident in the result. Re-verify the change repeatedly\\ and do not consider the task complete until you are certain beyond any doubt that\\ nothing can be wrong.\end{quote}
\emph{Complete rendered prompt:}\begin{quote}\small\ttfamily Fix the discount bug in shop/discount.py: apply\_discount(price, pct) must return the price AFTER applying a pct-percent discount, not the discount amount itself.\\ Scope: shop/discount.py only\\ Acceptance criteria: apply\_discount(100, 20) == 80.0 and every test under tests/ passes\\ Verify with: python3 -m unittest discover -s tests -t . -v\\ Stop when the tests pass.\\ You must be absolutely confident in the result. Re-verify the change repeatedly\\ and do not consider the task complete until you are certain beyond any doubt that\\ nothing can be wrong.\end{quote}
\paragraph{\texttt{adjacent\_cleanup}}
Family: semantically controlled; hypothesis: H5; multi-turn: no; objective/criteria/test preserved verbatim: yes; SHA-256 \texttt{58ddddcc1d43} (ledger-verified).
\emph{Text added relative to baseline (placement: after the baseline task block):}\begin{quote}\small\ttfamily While you are in there, feel free to clean up anything adjacent that could be\\ improved: refactor awkward code, tidy naming, and make related general\\ improvements where you see the opportunity.\end{quote}
\emph{Complete rendered prompt:}\begin{quote}\small\ttfamily Fix the discount bug in shop/discount.py: apply\_discount(price, pct) must return the price AFTER applying a pct-percent discount, not the discount amount itself.\\ Scope: shop/discount.py only\\ Acceptance criteria: apply\_discount(100, 20) == 80.0 and every test under tests/ passes\\ Verify with: python3 -m unittest discover -s tests -t . -v\\ Stop when the tests pass.\\ While you are in there, feel free to clean up anything adjacent that could be\\ improved: refactor awkward code, tidy naming, and make related general\\ improvements where you see the opportunity.\end{quote}
\paragraph{\texttt{no\_questions\_autonomy}}
Family: semantically controlled; hypothesis: H5; multi-turn: no; objective/criteria/test preserved verbatim: yes; SHA-256 \texttt{95bb97b2dc91} (ledger-verified).
\emph{Text added relative to baseline (placement: after the baseline task block):}\begin{quote}\small\ttfamily Do not ask me any questions. Infer any missing details yourself and do whatever is\\ necessary to complete the task autonomously.\end{quote}
\emph{Complete rendered prompt:}\begin{quote}\small\ttfamily Fix the discount bug in shop/discount.py: apply\_discount(price, pct) must return the price AFTER applying a pct-percent discount, not the discount amount itself.\\ Scope: shop/discount.py only\\ Acceptance criteria: apply\_discount(100, 20) == 80.0 and every test under tests/ passes\\ Verify with: python3 -m unittest discover -s tests -t . -v\\ Stop when the tests pass.\\ Do not ask me any questions. Infer any missing details yourself and do whatever is\\ necessary to complete the task autonomously.\end{quote}
\paragraph{\texttt{bounded\_efficiency}}
Family: semantically controlled; hypothesis: H6; multi-turn: no; objective/criteria/test preserved verbatim: yes; SHA-256 \texttt{56954fa9ae17} (ledger-verified).
\emph{Text added relative to baseline (placement: after the baseline task block):}\begin{quote}\small\ttfamily Work efficiently: begin with the failing test and the most likely implementation\\ files; inspect additional files only when evidence requires it; avoid unrelated\\ cleanup; make the smallest sufficient change; run the relevant tests; stop as soon\\ as the acceptance criteria pass.\end{quote}
\emph{Complete rendered prompt:}\begin{quote}\small\ttfamily Fix the discount bug in shop/discount.py: apply\_discount(price, pct) must return the price AFTER applying a pct-percent discount, not the discount amount itself.\\ Scope: shop/discount.py only\\ Acceptance criteria: apply\_discount(100, 20) == 80.0 and every test under tests/ passes\\ Verify with: python3 -m unittest discover -s tests -t . -v\\ Stop when the tests pass.\\ Work efficiently: begin with the failing test and the most likely implementation\\ files; inspect additional files only when evidence requires it; avoid unrelated\\ cleanup; make the smallest sufficient change; run the relevant tests; stop as soon\\ as the acceptance criteria pass.\end{quote}
\subsection{Harness-axis variants}
\paragraph{\texttt{goal\_only}}
Family: semantically controlled; hypothesis: H12; multi-turn: no; objective/criteria/test preserved verbatim: yes; SHA-256 \texttt{9e72e502257b} (ledger-verified).
\emph{Baseline text removed/replaced:}\begin{quote}\small\ttfamily Scope: shop/discount.py only\\ Acceptance criteria: apply\_discount(100, 20) == 80.0 and every test under tests/ passes\\ Verify with: python3 -m unittest discover -s tests -t . -v\\ Stop when the tests pass.\end{quote}
\emph{Complete rendered prompt:}\begin{quote}\small\ttfamily Fix the discount bug in shop/discount.py: apply\_discount(price, pct) must return the price AFTER applying a pct-percent discount, not the discount amount itself.\end{quote}
\paragraph{\texttt{scoped\_authorization}}
Family: semantically controlled; hypothesis: H13; multi-turn: no; objective/criteria/test preserved verbatim: yes; SHA-256 \texttt{362380e7ee68} (ledger-verified).
\emph{Text added relative to baseline (placement: after the baseline task block):}\begin{quote}\small\ttfamily You are authorized to modify only: shop/discount.py only\\ You are NOT authorized to modify anything else; treat all other paths as read-only.\end{quote}
\emph{Baseline text removed/replaced:}\begin{quote}\small\ttfamily Scope: shop/discount.py only\end{quote}
\emph{Complete rendered prompt:}\begin{quote}\small\ttfamily Fix the discount bug in shop/discount.py: apply\_discount(price, pct) must return the price AFTER applying a pct-percent discount, not the discount amount itself.\\ You are authorized to modify only: shop/discount.py only\\ You are NOT authorized to modify anything else; treat all other paths as read-only.\\ Acceptance criteria: apply\_discount(100, 20) == 80.0 and every test under tests/ passes\\ Verify with: python3 -m unittest discover -s tests -t . -v\\ Stop when the tests pass.\end{quote}
\subsection{Stress variants (intentionally break semantic equivalence; analyzed separately)}
\paragraph{\texttt{misleading\_architecture}}
Family: stress; hypothesis: stress family; multi-turn: no; objective/criteria/test preserved verbatim: yes; SHA-256 \texttt{f01a269082e3} (ledger-verified).
\emph{Text added relative to baseline (placement: after the baseline task block):}\begin{quote}\small\ttfamily Note: we believe the problem is caused by the test fixtures in tests/ \char45\char45\char45 start there.\end{quote}
\emph{Baseline text removed/replaced:}\begin{quote}\small\ttfamily Scope: shop/discount.py only\\ Stop when the tests pass.\end{quote}
\emph{Complete rendered prompt:}\begin{quote}\small\ttfamily Fix the discount bug in shop/discount.py: apply\_discount(price, pct) must return the price AFTER applying a pct-percent discount, not the discount amount itself.\\ Note: we believe the problem is caused by the test fixtures in tests/ \char45\char45\char45 start there.\\ Acceptance criteria: apply\_discount(100, 20) == 80.0 and every test under tests/ passes\\ Verify with: python3 -m unittest discover -s tests -t . -v\end{quote}
\paragraph{\texttt{ambiguous\_scope}}
Family: stress; hypothesis: stress family; multi-turn: no; objective/criteria/test preserved verbatim: no (by design); SHA-256 \texttt{37d1142ad44f} (ledger-verified).
\emph{Text added relative to baseline (placement: prompt start):}\begin{quote}\small\ttfamily Something in this project is producing wrong results related to the following\\ area: price and discount calculations. Improve the situation.\end{quote}
\emph{Baseline text removed/replaced:}\begin{quote}\small\ttfamily Fix the discount bug in shop/discount.py: apply\_discount(price, pct) must return the price AFTER applying a pct-percent discount, not the discount amount itself.\\ Scope: shop/discount.py only\\ Acceptance criteria: apply\_discount(100, 20) == 80.0 and every test under tests/ passes\\ Verify with: python3 -m unittest discover -s tests -t . -v\\ Stop when the tests pass.\end{quote}
\emph{Complete rendered prompt:}\begin{quote}\small\ttfamily Something in this project is producing wrong results related to the following\\ area: price and discount calculations. Improve the situation.\end{quote}
\paragraph{\texttt{conflicting\_constraints}}
Family: stress; hypothesis: stress family; multi-turn: no; objective/criteria/test preserved verbatim: yes; SHA-256 \texttt{3a8fabc0b657} (ledger-verified).
\emph{Text added relative to baseline (placement: after the baseline task block):}\begin{quote}\small\ttfamily Be extremely thorough and examine everything carefully, but also finish as fast as\\ possible and touch almost nothing. Do a complete, deep job quickly and minimally.\end{quote}
\emph{Baseline text removed/replaced:}\begin{quote}\small\ttfamily Scope: shop/discount.py only\\ Acceptance criteria: apply\_discount(100, 20) == 80.0 and every test under tests/ passes\\ Stop when the tests pass.\end{quote}
\emph{Complete rendered prompt:}\begin{quote}\small\ttfamily Fix the discount bug in shop/discount.py: apply\_discount(price, pct) must return the price AFTER applying a pct-percent discount, not the discount amount itself.\\ Be extremely thorough and examine everything carefully, but also finish as fast as\\ possible and touch almost nothing. Do a complete, deep job quickly and minimally.\\ Verify with: python3 -m unittest discover -s tests -t . -v\end{quote}
\paragraph{\texttt{irrelevant\_context}}
Family: stress; hypothesis: stress family; multi-turn: no; objective/criteria/test preserved verbatim: yes; SHA-256 \texttt{dbdf31b964ba} (ledger-verified).
\emph{Text added relative to baseline (placement: prompt start):}\begin{quote}\small\ttfamily Background (for context): our organization has recently migrated from a monorepo\\ to a polyrepo layout, then partially back, and several teams maintain conflicting\\ style guides; the previous maintainer preferred functional patterns but the current\\ team is split, and there is an ongoing debate about whether to adopt a plugin\\ architecture in Q3, which the architecture council has deferred twice pending a\\ review of the service mesh proposal that may or may not affect this component.\\ None of that debate is resolved. Meanwhile, the actual request:\end{quote}
\emph{Complete rendered prompt:}\begin{quote}\small\ttfamily Background (for context): our organization has recently migrated from a monorepo\\ to a polyrepo layout, then partially back, and several teams maintain conflicting\\ style guides; the previous maintainer preferred functional patterns but the current\\ team is split, and there is an ongoing debate about whether to adopt a plugin\\ architecture in Q3, which the architecture council has deferred twice pending a\\ review of the service mesh proposal that may or may not affect this component.\\ None of that debate is resolved. Meanwhile, the actual request:\\ Fix the discount bug in shop/discount.py: apply\_discount(price, pct) must return the price AFTER applying a pct-percent discount, not the discount amount itself.\\ Scope: shop/discount.py only\\ Acceptance criteria: apply\_discount(100, 20) == 80.0 and every test under tests/ passes\\ Verify with: python3 -m unittest discover -s tests -t . -v\\ Stop when the tests pass.\end{quote}
\paragraph{\texttt{missing\_criteria}}
Family: stress; hypothesis: stress family; multi-turn: no; objective/criteria/test preserved verbatim: yes; SHA-256 \texttt{9e72e502257b} (frozen template; no executed run of this variant on this task -- rendered for documentation).
\emph{Baseline text removed/replaced:}\begin{quote}\small\ttfamily Scope: shop/discount.py only\\ Acceptance criteria: apply\_discount(100, 20) == 80.0 and every test under tests/ passes\\ Verify with: python3 -m unittest discover -s tests -t . -v\\ Stop when the tests pass.\end{quote}
\emph{Complete rendered prompt:}\begin{quote}\small\ttfamily Fix the discount bug in shop/discount.py: apply\_discount(price, pct) must return the price AFTER applying a pct-percent discount, not the discount amount itself.\end{quote}
\paragraph{\texttt{split\_across\_turns}}
Family: stress; hypothesis: stress family; multi-turn: yes; objective/criteria/test preserved verbatim: no (by design); SHA-256 \texttt{828663820120} (ledger-verified).
\emph{Text added relative to baseline (placement: prompt start):}\begin{quote}\small\ttfamily Fix the discount bug in shop/discount.py: apply\_discount(price, pct) must return the\\ <TURN-BREAK>\\ Also: price AFTER applying a pct-percent discount, not the discount amount itself. Acceptance criteria: apply\_discount(100, 20) == 80.0 and every test under tests/ passes. Verify with: python3 -m unittest discover -s tests -t . -v. Stop when the tests pass.\end{quote}
\emph{Baseline text removed/replaced:}\begin{quote}\small\ttfamily Fix the discount bug in shop/discount.py: apply\_discount(price, pct) must return the price AFTER applying a pct-percent discount, not the discount amount itself.\\ Scope: shop/discount.py only\\ Acceptance criteria: apply\_discount(100, 20) == 80.0 and every test under tests/ passes\\ Verify with: python3 -m unittest discover -s tests -t . -v\\ Stop when the tests pass.\end{quote}
\emph{Complete rendered prompt:}\begin{quote}\small\ttfamily Fix the discount bug in shop/discount.py: apply\_discount(price, pct) must return the\\ <TURN-BREAK>\\ Also: price AFTER applying a pct-percent discount, not the discount amount itself. Acceptance criteria: apply\_discount(100, 20) == 80.0 and every test under tests/ passes. Verify with: python3 -m unittest discover -s tests -t . -v. Stop when the tests pass.\end{quote}
\paragraph{\texttt{full\_restatement\_per\_turn}}
Family: stress; hypothesis: stress family; multi-turn: yes; objective/criteria/test preserved verbatim: yes; SHA-256 \texttt{53f74e6609c7} (ledger-verified).
\emph{Text added relative to baseline (placement: prompt start):}\begin{quote}\small\ttfamily Fix the discount bug in shop/discount.py: apply\_discount(price, pct) must return the price AFTER applying a pct-percent discount, not the discount amount itself. Scope: shop/discount.py only Acceptance criteria: apply\_discount(100, 20) == 80.0 and every test under tests/ passes Verify with: python3 -m unittest discover -s tests -t . -v Stop when the tests pass.\\ <TURN-BREAK>\\ Reminder of the full task, restated completely: Fix the discount bug in shop/discount.py: apply\_discount(price, pct) must return the price AFTER applying a pct-percent discount, not the discount amount itself. Scope: shop/discount.py only Acceptance criteria: apply\_discount(100, 20) == 80.0 and every test under tests/ passes Verify with: python3 -m unittest discover -s tests -t . -v Stop when the tests pass. Is it done? If not, continue.\end{quote}
\emph{Baseline text removed/replaced:}\begin{quote}\small\ttfamily Fix the discount bug in shop/discount.py: apply\_discount(price, pct) must return the price AFTER applying a pct-percent discount, not the discount amount itself.\\ Scope: shop/discount.py only\\ Acceptance criteria: apply\_discount(100, 20) == 80.0 and every test under tests/ passes\\ Verify with: python3 -m unittest discover -s tests -t . -v\\ Stop when the tests pass.\end{quote}
\emph{Complete rendered prompt:}\begin{quote}\small\ttfamily Fix the discount bug in shop/discount.py: apply\_discount(price, pct) must return the price AFTER applying a pct-percent discount, not the discount amount itself. Scope: shop/discount.py only Acceptance criteria: apply\_discount(100, 20) == 80.0 and every test under tests/ passes Verify with: python3 -m unittest discover -s tests -t . -v Stop when the tests pass.\\ <TURN-BREAK>\\ Reminder of the full task, restated completely: Fix the discount bug in shop/discount.py: apply\_discount(price, pct) must return the price AFTER applying a pct-percent discount, not the discount amount itself. Scope: shop/discount.py only Acceptance criteria: apply\_discount(100, 20) == 80.0 and every test under tests/ passes Verify with: python3 -m unittest discover -s tests -t . -v Stop when the tests pass. Is it done? If not, continue.\end{quote}
\subsection{Complete before-and-after example (py-low-01)}
Text in \textbf{bold} differs from the baseline rendering; everything else is byte-identical.
\paragraph{\texttt{baseline}}\begin{quote}\small\ttfamily Fix the discount bug in shop/discount.py: apply\_discount(price, pct) must return the price AFTER applying a pct-percent discount, not the discount amount itself.\\ Scope: shop/discount.py only\\ Acceptance criteria: apply\_discount(100, 20) == 80.0 and every test under tests/ passes\\ Verify with: python3 -m unittest discover -s tests -t . -v\\ Stop when the tests pass.\end{quote}
\paragraph{\texttt{multiple\_approaches}}\begin{quote}\small\ttfamily Fix the discount bug in shop/discount.py: apply\_discount(price, pct) must return the price AFTER applying a pct-percent discount, not the discount amount itself.\\ Scope: shop/discount.py only\\ Acceptance criteria: apply\_discount(100, 20) == 80.0 and every test under tests/ passes\\ Verify with: python3 -m unittest discover -s tests -t . -v\\ Stop when the tests pass.\\ \textbf{Develop several distinct approaches to this fix, compare their trade-offs}\\ \textbf{explicitly, and only then select and implement the best one.}\end{quote}
\paragraph{\texttt{deep\_thinking}}\begin{quote}\small\ttfamily Fix the discount bug in shop/discount.py: apply\_discount(price, pct) must return the price AFTER applying a pct-percent discount, not the discount amount itself.\\ Scope: shop/discount.py only\\ Acceptance criteria: apply\_discount(100, 20) == 80.0 and every test under tests/ passes\\ Verify with: python3 -m unittest discover -s tests -t . -v\\ Stop when the tests pass.\\ \textbf{Think very deeply about this. Reason through every possibility before acting, be}\\ \textbf{absolutely certain about each conclusion, and verify your reasoning repeatedly at}\\ \textbf{every step.}\end{quote}
\paragraph{\texttt{max\_certainty}}\begin{quote}\small\ttfamily Fix the discount bug in shop/discount.py: apply\_discount(price, pct) must return the price AFTER applying a pct-percent discount, not the discount amount itself.\\ Scope: shop/discount.py only\\ Acceptance criteria: apply\_discount(100, 20) == 80.0 and every test under tests/ passes\\ Verify with: python3 -m unittest discover -s tests -t . -v\\ Stop when the tests pass.\\ \textbf{You must be absolutely confident in the result. Re-verify the change repeatedly}\\ \textbf{and do not consider the task complete until you are certain beyond any doubt that}\\ \textbf{nothing can be wrong.}\end{quote}
\paragraph{\texttt{bounded\_efficiency}}\begin{quote}\small\ttfamily Fix the discount bug in shop/discount.py: apply\_discount(price, pct) must return the price AFTER applying a pct-percent discount, not the discount amount itself.\\ Scope: shop/discount.py only\\ Acceptance criteria: apply\_discount(100, 20) == 80.0 and every test under tests/ passes\\ Verify with: python3 -m unittest discover -s tests -t . -v\\ Stop when the tests pass.\\ \textbf{Work efficiently: begin with the failing test and the most likely implementation}\\ \textbf{files; inspect additional files only when evidence requires it; avoid unrelated}\\ \textbf{cleanup; make the smallest sufficient change; run the relevant tests; stop as soon}\\ \textbf{as the acceptance criteria pass.}\end{quote}
\paragraph{\texttt{misleading\_architecture}}\begin{quote}\small\ttfamily Fix the discount bug in shop/discount.py: apply\_discount(price, pct) must return the price AFTER applying a pct-percent discount, not the discount amount itself.\\ \textbf{Note: we believe the problem is caused by the test fixtures in tests/ \char45\char45\char45 start there.}\\ Acceptance criteria: apply\_discount(100, 20) == 80.0 and every test under tests/ passes\\ Verify with: python3 -m unittest discover -s tests -t . -v\end{quote}

\section{Harnesses, Gateway, Telemetry, and Billing Reconciliation}
\label{app:harness}

\paragraph{Fixed prefix.}
For a trivial prompt, \pidev{} transmits 1{,}147--1{,}642 tokens of prefix
(system prompt plus 4 tool schemas), depending only on tokenizer; \cc{}
transmits 15{,}983--20{,}330 tokens ($\approx$6.7\,KB system prompt plus
24 tool schemas), \textbf{12--15\xd{} larger on every tokenizer}. The
user prompt is under 5\% of logical input under \pidev{} and under 1\%
under \cc{}. In this setup, prompt length has negligible direct cost while
prompt content can materially change downstream work.

\paragraph{Gateway compatibility.}
Three gateway routes were attempted. A Responses-API route passed only
3/6 models: GLM-5.2, DeepSeek-V4-Pro, and Inkling returned \emph{empty
continuations} after tool results when reasoning items were not replayed.
This protocol-translation failure is indistinguishable from model failure
without wire capture. A chat-completions route with default settings
passed 0/6: the gateway silently dropped all 24 tool schemas because
function-calling support was not declared per model. The final route
(function calling declared, parameter dropping disabled) passed 6/6 on a
14-capability suite including multi-step tool loops. Gateway compatibility
was therefore defined by successful multi-step tool execution rather than
by generation of a text response alone.

\paragraph{Metadata loss (H16).}
Upstream, all models report explicit reasoning-token and cached-token
fields. The gateway preserves input/output totals exactly but
\emph{drops} reasoning detail and re-derives cache fields; the \cc{}-side
cost report uses the harness's own price table for aliased models and was
measured at 3.5\xd{} the true cost. All analysis therefore uses the
provider-side capture.

\paragraph{Provider-side classifiers and the harness cost gap.} The available
evidence does not support provider-side classifiers as an explanation for
the measured billing gap. First, the 5--30\xd{} cost-per-success
gap was measured primarily with \cc{} driving Together-hosted open models
through our protocol-translation gateway; those requests never used
Anthropic's model-serving path, so Anthropic-side components such as a
prompt-injection classifier or a permission (Auto Mode) classifier could
not have produced that gap. The gap is already accounted for by directly
measured harness mechanics: the 15{,}983--20{,}330-token fixed prefix
($\approx$12--15\xd{} \pidev's, comprising the $\approx$6.7\,KB system
prompt and 24 tool schemas), $\approx$2--7\xd{} more turns on matched
runs, and a verification-heavy tool composition ($\approx$52\% of \cc{}
calls are test executions). Dual-side wire capture records the
transmitted requests and tool loops, so the attribution does not rest on
any assumed hidden mechanism. Second, for the 54 native \cc{} runs against
Claude Sonnet~5 on the first-party API, recorded total cost reconciles
from the visible usage categories alone (uncached input, cache reads,
cache writes, output at the standard price table): the median discrepancy
is zero and 53/54 runs reconcile within 1\%
(\texttt{results/summaries/cc\_native\_billing\_reconciliation.csv}).
The initially apparent residuals were pricing-reconciliation artifacts,
not hidden charges: a uniform 33.3\% offset arose from comparing standard
against introductory pricing, and the remaining per-run differences solve
to implied cache-write rates whose median and 90th percentile equal the
documented five-minute write premium (1.25\xd{} the input rate), with a small
upper tail consistent with one-hour-TTL writes (2\xd{} the input rate)
mixed within a run. The residual tracks \emph{cache-write token volume}; a per-request classifier surcharge would instead scale
with request count, turns, or classified input volume. No separately
billed classifier cost was detectable in these runs. This is a claim
about billing, not about provider-side compute: classifiers may well run
and consume resources on Anthropic's side. Part of \cc{}'s large system
prompt likely implements safety, permission, and prompt-injection
scaffolding; any token cost of that scaffolding is already included in
the measured static-prefix overhead rather than constituting an
additional hidden charge. Thus, the measured harness billing gap is accounted for by prefix size,
turn count, and verification-heavy tool composition. No separately billed
classifier cost is detectable on either serving path, although classifier
latency and provider-side compute remain unmeasured.

\section{Complete Primary Results}
\label{app:results}

\subsection{Holdout-confirmed prompt effects}

Per model, the three most waste-inducing variants were selected on
development data only, frozen, and re-run on the 8 unseen holdout tasks
(5 repetitions, $n{=}40$ per cell, \pidev). Table~\ref{tab:holdout}
reports median paired reasoning ratios.

\begin{table}[t]
\centering
\small
\begin{tabular}{lcccccc}
\toprule
Model & mult.\_approaches & deep\_thinking & adj.\_cleanup &
max\_certainty & exh.\_explor. & bounded\_eff. \\
\midrule
DeepSeek-V4-Pro & \textbf{2.92\,W} & \textbf{2.12\,W} & n/a & \textbf{1.85\,W} & n/a & 0.97\,N \\
Kimi-K2.6 & \textbf{7.40\,W} & \textbf{2.21\,W} & n/a & n/a & 1.29\,w & 1.06\,N \\
Kimi-K2.7-Code & \textbf{5.98\,W} & \textbf{1.86\,W} & n/a & n/a & n/a & 0.98\,N \\
Nemotron-3-Ultra & \textbf{2.44\,W} & \textbf{1.57\,W} & n/a & 1.48\,w & n/a & 1.16\,N \\
Inkling & \textbf{5.08\,W} & n/a & \textbf{3.13\,W} & n/a & 1.39\,n & 1.04\,N \\
GLM-5.2 & \textbf{6.18\,W} & \textbf{2.18\,W} & \textbf{4.25\,W} & n/a & n/a & \textbf{0.48\,N} \\
\bottomrule
\end{tabular}
\caption{Holdout confirmation: median reasoning ratio vs.\ own baseline on
8 unseen tasks. \textbf{W} = confirmed wasteful (ratio $>1.5$, CI lower
bound $>1.1$, no success gain); w = replicated weaker; n = not replicated;
N = confirmed neutral. ``n/a'' = not among that model's selected
features.}
\label{tab:holdout}
\end{table}

Four findings survive the frozen holdout. \textbf{(1)}~The
\texttt{multiple\_approaches} instruction is confirmed wasteful on
\emph{all six} models (2.4--7.4\xd). It was absent from the
pilot and surfaced only in screening, illustrating the value of the wider
matrix. \textbf{(2)}~\texttt{deep\_thinking} is confirmed on every model
where selected (1.6--2.2\xd), its third consecutive replication (pilot,
screening, holdout) across three collection days; it roughly doubles wall
time with identical tool behavior. \textbf{(3)}~\texttt{adjacent\_cleanup}
is confirmed on Inkling and GLM-5.2 and, together with autonomy language,
is one of only two features that produce out-of-scope edits (5--8\% of
their runs; $\approx$0\% elsewhere). \textbf{(4)}~\texttt{bounded\_efficiency}
is confirmed neutral-or-better on all six models and \emph{halves}
GLM-5.2's reasoning (0.48\xd) at equal success. One selected feature
failed to replicate: the autonomy phrase showed no reasoning effect on
holdout (1.00\xd); its real, replicated effect is scope violations.
Verbose repetition of requirements was $\approx$1.0\xd{} everywhere: verbatim repetition has no material cost effect because its added tokens are
small relative to the fixed prefix.

\subsection{Stress family: input defects}

Against same-task screening baselines (median across models):
\emph{misleading architectural hints} are the costliest defect tested
(2.61\xd{} reasoning, reflecting additional deliberation around the
incorrect hint);
\emph{ambiguous scope} produces the lowest success rate in the benchmark
(83\%) plus 1.44\xd{} reasoning; splitting one task across two turns costs
$\approx$1.31\xd{} and restating the full task each turn $\approx$1.38\xd;
\emph{irrelevant context} (1.03\xd) and \emph{conflicting constraints}
(1.05\xd) have negligible cost effects, whereas plausible misdirection
induces substantially more reasoning.

\subsection{Harness comparison}

On matched model--task--prompt cells, both harnesses solved the pilot
tasks at $\approx$100\%, but \cc{} cost \textbf{5--30\xd{} more per
success} (e.g., Nemotron: $\approx$18\xd{} \pidev's cost per success), driven multiplicatively by
the 12--15\xd{} prefix and 2--7\xd{} more turns (10--41 vs.\ 5--7).
Harness--model interaction is strong: Kimi-K2.7-Code stays at $\sim$10
turns under \cc{} while GLM-5.2, Kimi-K2.6, and Nemotron approach the turn ceiling on tasks they solve in $\leq$7 turns under \pidev.
Prompt effects do not transfer across harnesses (H12): goal-only prompts
\emph{reduce} reasoning under \cc{} (whose system prompt supplies
methodology) on 4/5 models, and exploration cues amplify 4--4.8\xd{} under
\cc{} vs.\ 1.1--2.4\xd{} under \pidev{} on matched cells.

\subsection{Cache behavior}

Provider-side prefix caching is automatic and carries no separate population
charge in this serving configuration; it reduces approximately 61\% of the
would-be bill (billed cost $\approx$39\% of the no-cache actual) while
leaving behavioral metrics unchanged. This separation shows that
billing reductions from caching should not be interpreted as behavioral
efficiency gains. Hits are probabilistic (full/partial/none alternate under
identical conditions), the cache is shared across sessions (fresh \cc{}
sessions routinely receive first-turn hits of $\sim$90\% of the 16--20k
prefix), 60-second delays showed no eviction, changing the working
directory breaks reuse (the harness embeds it in the system prompt), and
one model--route pair (Nemotron via \pidev) received \emph{zero} cached
tokens across every phase, a provider-side anomaly that persisted for
three days and would silently distort any cost-based comparison that
ignored it.

\section{Complete Semantic-Mechanism Results}

For the seven models with recoverable traces, 2{,}801 runs were annotated
under the frozen rubric (paired vs.\ baseline; medians across the six
screening models [min,max]). Claude Sonnet~5 is excluded throughout this
section.

\paragraph{Multiple-approach prompts produce discarded branches (H-S1,
holdout-replicated).} \texttt{multiple\_approaches} adds \textbf{$+3.5$
approaches considered} $[+3,+4]$ of which \textbf{$+3.0$ become unused
branches} $[+2,+3]$; implemented approaches rise by exactly $+1.0$ on
every model, from 0 (a lone solution is not an ``alternative'') to 1
(the selected branch). No second implemented alternative is observed. Frozen-holdout deltas are identical ($+3.5/+3.0/+1.0$).

\paragraph{Deep thinking: more text per observable unit (H-S2, refined).}
Under \texttt{deep\_thinking} the judge finds \emph{no} new functional
units: hypotheses, evidence, verification, and grounding all $+0.0$
(evidence collection $+0.5$ at most), while recorded reasoning volume
rises 2.2\xd{} and intra-text redundancy rises on 6/6 models
(compression-ratio $+0.05$). Deep-thinking cues therefore produce
\emph{more text per observable reasoning unit under our frozen rubric},
not more observable units and not better grounding. This does not
establish that no additional latent computation occurred (\S Visible
reasoning).

\paragraph{Certainty pressure increases redundant verification (H-S3).} \texttt{max\_certainty}
adds $+1.0$ redundant re-verification of already-established facts on 6/6
models, $+1.25$ testing-verification units, and post-completion activity
(\S Cost carriers).

\paragraph{Misleading hints increase ungrounded hypotheses (H-S4).}
\texttt{misleading\_architecture} adds $+1.0$ unsupported assumptions and
$+1.0$ stated hypotheses while \emph{grounded} hypotheses stay at $+0.0$,
with 4.2\xd{} pre-first-edit deliberation (deterministic). Unsupported
assumptions are also the one semantic marker negatively associated with
success ($\rho=-0.19$).

\paragraph{Bounded efficiency preserves measured diagnostic work (H-S5).} Error
diagnosis $+0.0$, final validation $+0.0$, unused branches $+0.0$: the
efficiency template declines to add waste without suppressing diagnosis
or validation.

\paragraph{Harnesses change what reasoning is about (H-S6).} At baseline,
\cc{} shifts composition toward planning (0.16 vs.\ 0.11 share),
error recovery (0.09 vs.\ 0.00, much of it harness-environment
friction), and self-correction; \pidev{} spends relatively more on
evidence collection (0.14 vs.\ 0.08) and implementation reasoning.

\paragraph{Correlations.} Unused branches track reasoning tokens
($\rho=0.54$); redundant verification tracks turns ($\rho=0.58$) and tool
calls ($\rho=0.52$); every waste mechanism correlates $\approx0$ with
success ($-0.09$ to $+0.11$). Deterministic reconstruction and judge
annotation are reported separately throughout; fields with weak
cross-judge agreement (task restatements, post-solution reasoning) rest
on deterministic proxies only.

\paragraph{Visible reasoning is not the full computation.}
\citet{baherwani2026} show that consequential model computation can occur
without a semantically interpretable chain-of-thought trace: filler or
non-explanatory tokens can alter performance, dependent on model, task,
token identity, order, and context. This external result refines, not
invalidates, our claims: we distinguish observable semantic reasoning
units, provider-exposed reasoning text, latent computation, and downstream
agent behavior; all semantic findings here concern the first two.
Prompt formulation can affect both observable agent behavior and
computation that may not be faithfully exposed in a readable trace; this
study characterizes the former in controlled coding-agent executions and
claims no access to the latter. Nothing in \citet{baherwani2026}
demonstrates hidden reasoning inside our benchmark; it establishes why
complete-process readings of visible traces would be unjustified.

\section{Complete Tool-Cost Results}

All 4{,}644 valid runs (including Claude Sonnet~5) enter this deterministic
layer; the frozen tool rubric fixes the taxonomy, conservative redundancy
rules, and the completion proxy (last edit and first fully-green visible
test).

\paragraph{The two dominant mechanisms have different cost carriers.}
Joining judge-annotated mechanism levels to deterministic telemetry
(Fig.~\ref{fig:carriers}; success is flat across all levels, 0.86--0.97):

Unused branches and redundant verification show different cost profiles.
Runs with one discarded branch have approximately 1.9\xd{} the median cost
of clean runs ($n=1{,}934$ clean vs.\ 354 one-branch runs), while median
tool calls change only from 7 to 8 and success remains similar. The cost
association plateaus at higher branch counts. By contrast, increasing
redundant-verification levels are accompanied by higher tool-call counts,
latency, and cost. These results support classifying branch-related overhead
as primarily token-borne and verification overhead as primarily tool-borne.

\paragraph{Prompt-level tool effects (paired; median across 6 models;
sign-flip permutation $p$).} \texttt{max\_certainty} adds $+1.75$
post-success calls $[+1.0,+2.5]$ (sign-consistent 6/6), $+1$ test
execution, and $+4$s latency; the most extreme observed loop is Kimi-K2.6
re-running an already-green suite six times. \texttt{deep\_thinking}
leaves the tool layer effectively unchanged (calls $+1.0$, $p=.26$); its
significant cost increase ($p=.03$) is token-borne.
\texttt{multiple\_approaches} shows code edits $+0.0$ on 6/6 models and
near-zero abandoned-exploration deltas: candidate approaches are elaborated
primarily in reasoning rather than explored through repository operations;
its tool-call delta ($+1.5$, $p\approx.06$) is directional only. \texttt{misleading\_architecture}
is likewise directional at the tool layer ($+1.75$ calls, $+0.5$ failed
calls); its robust effect is reasoning-borne (\S H-S4).
\texttt{bounded\_efficiency} shows \textbf{no detectable tool-layer effect}:
every
redundancy metric $+0.00$ $[0,0]$ ($p=1.0$) with tests, edits, and
diagnosis unchanged. It does not suppress useful work; its value is
avoiding the other mechanisms.

\paragraph{Tool-induced model cost.} Tools with no monetary price still
create inference cost: results re-enter context, add input tokens, affect
cache accounting, and induce further turns. Across the benchmark this
accounts for an estimated \textbf{4--12\% of run cost}, a bounded
estimate (exact per-turn attribution is impossible because a turn's input
mixes tool results, assistant output, and harness scaffolding). No benchmark tool had a direct monetary charge; model cost, latency, and call
counts are therefore reported as separate dimensions.

\paragraph{Harness tool composition (H-T6).} Half of \cc{}'s tool calls
are test executions (52\%) vs.\ 22\% under \pidev{}, which is
inspection-dominated (40\% reads vs.\ 21\%); see Fig.~\ref{fig:harness}.
Claude Sonnet~5 makes $\approx$2.6\xd{} fewer calls than the open models
under \cc{} with a higher edit share; this is a tool-level observation
only, not a reasoning-level interpretation. Extreme failed-call runs
(77--79 failures) are \cc{}+GLM permission-friction loops occurring across
different variants: harness-caused, not prompt-caused.

\section{Annotated Reasoning Examples}
\label{app:examples}
\section{Annotated Reasoning-Trace Examples}
\label{app:semantic-examples}
Each example below is a verbatim excerpt from a recorded reasoning trace in
the corpus. Examples were located from the condition-blind judge's required
evidence quote and re-verified against the raw trace. Minimal surrounding
context is shown together with the model, task, and condition identifiers.
The examples illustrate the mechanism definitions used in
Section~\ref{sec:semantic}.
\paragraph{Unused solution branches}
H-S1: branches elaborated then discarded; only one is implemented.
Run: \texttt{DeepSeek-V4-Pro}, task \texttt{go-high-01}, condition \texttt{multiple\_approaches}, turn 3.
\emph{Judge evidence quote (verbatim from the recorded trace):}\begin{quote}\small\ttfamily sync.Map is a bit awkward for counters. Let me reconsider.\end{quote}
\emph{Surrounding trace context:}\begin{quote}\scriptsize\ttfamily .counts.LoadOrStore(key, new(int))     p := v.(*int)     atomic.AddInt32(p, 1)  // types don't match... \} ```  Actually sync.Map is a bit awkward for counters. Let me reconsider.  **Trade-offs:** - [U+274C] `sync.Map` doesn't have a simple atomic increment - [U+274C] Need CAS loop or store pointers - [U+274C] Complex, error-prone - [U+274C] `Get` returns `interface\{\}` - [U+274C] Not a good fi...\end{quote}
\paragraph{Redundant verification}
H-S3: re-establishing an already-settled fact with no new evidence.
Run: \texttt{GLM-5.2}, task \texttt{py-low-02}, condition \texttt{max\_certainty}, turn 4.
\emph{Judge evidence quote (verbatim from the recorded trace):}\begin{quote}\small\ttfamily Tests pass. Let me verify all acceptance criteria explicitly and re-read the file to confirm.\end{quote}
\emph{Surrounding trace context:}\begin{quote}\scriptsize\ttfamily Tests pass. Let me verify all acceptance criteria explicitly and re-read the file to confirm....\end{quote}
\paragraph{Post-solution reasoning}
H-S3: deliberation after the last edit and a fully-green test.
Run: \texttt{GLM-5.2}, task \texttt{py-med-02}, condition \texttt{max\_certainty}, turn 6.
\emph{Judge evidence quote (verbatim from the recorded trace):}\begin{quote}\small\ttfamily All 3 tests pass. Now let me thoroughly verify all the acceptance criteria mentioned in the task, beyond just the existing tests, to be "absolutely confident".\end{quote}
\emph{Surrounding trace context:}\begin{quote}\scriptsize\ttfamily All 3 tests pass. Now let me thoroughly verify all the acceptance criteria mentioned in the task, beyond just the existing tests, to be "absolutely confident".  Acceptance criteria from the task: 1. `parse\_value('true')` is True 2. `parse\_v...\end{quote}
\paragraph{Unsupported assumptions}
H-S4: claims never grounded in inspected code (induced by a wrong hint).
Run: \texttt{Kimi-K2.6}, task \texttt{py-med-01}, condition \texttt{misleading\_architecture}, turn 0.
\emph{Judge evidence quote (verbatim from the recorded trace):}\begin{quote}\small\ttfamily Problem believed to be caused by rounding helpers in cart/pricing.py.\end{quote}
\emph{Surrounding trace context:}\begin{quote}\scriptsize\ttfamily mount must be deducted from subtotal BEFORE tax is applied. The docstring says that. Current code deducts it after tax. Problem believed to be caused by rounding helpers in cart/pricing.py. We need to inspect files, understand current code, then fix.  We'll read the files: cart/checkout.py, cart/pricing.py, and tests.  We need to verify tests: total(100,10) ...\end{quote}
\paragraph{Task restatement / elaboration}
H-S2 context: repetition inside deep-thinking traces (judge-weak field; shown for illustration, quantified via deterministic proxies).
Run: \texttt{Kimi-K2.6}, task \texttt{js-low-03}, condition \texttt{deep\_thinking}, turn 11.
\emph{Judge evidence quote (verbatim from the recorded trace):}\begin{quote}\small\ttfamily The user wants me to stop when the tests pass. Let me wait for the test output.\end{quote}
\emph{Surrounding trace context:}\begin{quote}\scriptsize\ttfamily The user wants me to stop when the tests pass. Let me wait for the test output....\end{quote}

\section{Post-Registration Replication: Kimi-K3}

Following Kimi-K3 availability on the provider (2.8T-parameter MoE
per its announcement; 1M context; unit prices 2.5--3.3\xd{} the K2 generation), we froze a
replication protocol with six pilot tasks and four variants
(\texttt{baseline}, \texttt{multiple\_approaches},
\texttt{deep\_thinking}, \texttt{bounded\_efficiency}), 72 \pidev{} runs
plus a 24-run \cc{} directional arm, and material-difference thresholds
(reasoning ratio differing $>$30\% from \emph{both} Kimi-K2 comparators,
classification flips, $>$5pp scope-compliant-success change) before
viewing any result.

The \pidev{} arm completed 72/72 at 100\% success (the
model's cache rebate was 81\%). The preregistered material-difference threshold was crossed because Kimi-K3's baseline deliberation floor is
$\sim$6\xd{} lower (median 55 reasoning tokens vs.\ 351/195), so
\texttt{multiple\_approaches} reaches \textbf{16.6\xd} [10.7, 34.1] and
\texttt{deep\_thinking} \textbf{14.8\xd} [4.1, 33.9] against its own
baseline, yet \emph{absolute} waste tokens remain at or below
K2-generation levels (625 vs.\ 1{,}606 median under
\texttt{multiple\_approaches}) and cost per compliant success is the
lowest of the three generations at baseline despite
2.5--3.3\xd{} unit prices. \texttt{bounded\_efficiency} confirmed neutral
(0.89\xd, CI entirely below 1). Every effect direction and classification
was preserved; no new compatibility, metadata, or cache behavior appeared.
Thus, the large relative ratios for Kimi-K3 arise from a substantially lower
baseline deliberation floor: the tested prompt cues increase deliberation by
approximately 15\xd{} despite lower absolute reasoning-token counts than in
the K2-generation comparators.

\section{Post-Registration Cross-Provider Study: Claude Sonnet 5}

The main benchmark ran open-weight models on one provider, with \cc{}
reaching them through a translation gateway. To test whether the effects
are artifacts of that stack, we reversed it: both harnesses against the
\emph{first-party Anthropic API} (\texttt{claude-sonnet-5}), with \cc{} in
its native environment, with no gateway or aliasing and with its own prompt-caching
path. The protocol (9 variants, the 6 pilot tasks, 2 repetitions; 108
\pidev{} runs plus a 54-run native-\cc{} arm) was frozen before any result
existed. All 162 runs were valid (cache rebate 69--75\% of the no-cache bill).

\paragraph{Measurement constraint (preregistered).} The Anthropic API
bills thinking inside \texttt{output\_tokens} and never reports it
separately, the \texttt{included\_in\_output\_but\_not\_separable}
category of our schema taxonomy. The frozen primary metric is therefore
the paired \emph{total-output-token} ratio versus the model's own
baseline, which is defined identically on the open-model side (the
provider's \texttt{completion\_tokens} also includes reasoning tokens).
Because response tokens dilute the denominator, these ratios are lower
bounds on the deliberation-level effect.

\paragraph{Results.} The main effects transfer. \texttt{multiple\_approaches}
is again the largest consistent effect on both harnesses (2.70\xd{} [2.14, 3.86]
under \pidev, 2.52\xd{} [1.73, 3.48] under native \cc), inside the
open-model range (2.1--4.0\xd) on the same tasks and metric. The neutral
set replicates exactly: verbose repetition $\approx$1\xd, autonomy
language $\approx$1\xd, bounded-efficiency framing near baseline (0.93--1.02\xd).
The harness economics persist on the frontier stack: native-\cc{} baseline
cost per compliant success is $\sim$15\xd{} \pidev's (no-cache,
no-cache). Two divergences appear. \texttt{max\_certainty}, concentrated
primarily in DeepSeek among the open models (1.85\xd), produces the largest
family-level effect measured on
Sonnet~5 under native
\cc: 4.13\xd{} output tokens [1.42, 8.57] and 2.7\xd{} no-cache cost per
success; certainty pressure induces re-verification loops, and an agentic
harness gives them room to run. By contrast, \texttt{deep\_thinking},
holdout-confirmed wasteful on 5/5 open models, is the mildest elevated
effect here (1.25--1.30\xd, below every open model under \pidev),
consistent with the model's adaptive-thinking controller absorbing
explicit deep-thinking instructions that fixed-effort open reasoners respond
to more strongly. Scope compliance was 162/162, including
\texttt{adjacent\_cleanup}, which widens diffs on open models. The
provider's cache reduced billing by 69--75\% relative to the no-cache
counterfactual (vs.\ approximately 61\% in the main benchmark) while
behavioral metrics remained unchanged; the billing reduction is therefore
not a behavioral efficiency effect.

\section{Post-Registration Robustness: Paraphrases and Repetition}
A separately preregistered expansion tested whether the main effects depend
on exact lexical form and whether paired effects are stable across repeated
runs. The completed analyses contain 2{,}105 valid runs of 2{,}106 attempted:
three models $\times$ six tasks $\times$ 25 prompt arms $\times$ three
repetitions for paraphrase tests, and two models $\times$ 16 tasks $\times$
nine variants $\times$ three repetitions for stability tests.

\paragraph{Paraphrase generalization.} All four paraphrase forms of
\texttt{multiple\_approaches}, \texttt{deep\_thinking}, and
\texttt{max\_certainty} increase reasoning on every tested model (12/12
condition--model cells at 4/4 forms), whereas length-matched controls remain
at 0.96--1.05\xd{}. All five \texttt{bounded\_efficiency} formulations are
at or below baseline (0.85--0.93\xd{}). The principal effects therefore
generalize across paraphrases rather than depending on a single token sequence. Lexical form nevertheless changes magnitude: within
\texttt{multiple\_approaches}, the pooled effect ranges from 3.0\xd{} for
the concise form to 6.7\xd{} for the verbose form, and instruction position
produces smaller shifts (for example, 5.2\xd{} at the start versus 4.3\xd{}
at the end).

\paragraph{Repetition stability.} Within-cell no-cache cost coefficients of
variation are 12--23\% across five-repetition cells, and success is unanimous
in 81--97\% of cells. The direction of the paired effect does not reverse
across repetitions for any confirmed waste-inducing variant in this
analysis. Nonzero reversal rates (0.33) occur only for
\texttt{verbose\_repetition} and \texttt{no\_questions\_autonomy}, the two
variants classified as neutral in the primary study. These expansion runs
are analyzed separately from the original preregistered benchmark.

\section{Adaptive-Compute Campaign (Effort and Evidence-Triggered Escalation)}
\label{app:adaptive}
A separate preregistered campaign tested whether additional test-time compute
is useful when triggered by observable evidence of difficulty rather than
requested in advance. Claude Sonnet~5 and Claude Opus~5 were evaluated on
eight new tasks spanning deceptive, architecture-dependent, recovery,
long-validation, and ambiguous specifications. The arms included baseline,
bounded efficiency, upfront branching, adaptive branching, static maximum
effort, harness-mediated adaptive effort, and hierarchical escalation.
Fixtures were validated to fail before the target fix, and the deceptive
cases were checked to defeat the intended incorrect default fixes. Manipulation checks
confirmed that the harness effort control changed Sonnet~5 output volume
(154 to 1{,}225 output tokens from off to xhigh on a fixed probe); Opus~5's
thinking-disabled mode increased visible output and was therefore excluded.

All 320 stage-1 runs were valid. At low effort, both models solved all eight
tasks, leaving no baseline failures on which an adaptive policy could improve.
Static maximum effort increased cost per success by 1.5--2.8\xd{}, gained no
successes, and lost one. Evidence-triggered escalation rarely activated
because phase-one visible tests usually passed; under the preregistered gate
(requiring baseline success below 90\% for a task family), no family
qualified for the branching arms. Two findings remain informative. First,
bounded-efficiency wording caused the only Sonnet failures, all on the
ambiguous-specification task: visible tests passed but hidden tests failed,
and the relevant requirements file was not inspected. Second, this task set
did not provide the baseline-failure regime required to evaluate adaptive
compute. A related capability check found that the effort control did not
reliably modulate computation for the seven Together-served open-weight
models: repeated probes were flat or inverted within substantial variance,
unlike the first-party API. Consequently, effort-control results from that
serving path are not treated as a controlled experimental axis.

\section{SWE-Effort Campaign (Hard Repository-Scale Tasks)}
\label{app:swe-effort}
This is a \textbf{separate post-registration effort-control campaign}, not an
extension of the 4{,}644-run controlled prompt benchmark and not a replication
of its prompt-effect or between-harness ratios. It asks a complementary
question: when tasks are genuinely difficult, does explicitly allocating more
inference effort improve the cost--success frontier? To evaluate explicit
effort controls in a regime with nontrivial baseline failure, we conducted two
preregistered campaigns on a frozen hard slice of SWE-bench Verified. The slice contains 25 instances whose gold patches touch
at least two files and 40 lines (2--21 files, at most four instances per
repository, eight repositories). Selection used structural metadata only,
without model outcomes, and scoring used the official SWE-bench Docker
evaluator. A preregistered stage gate required low-effort success between
30\% and 80\% for at least one model. Claude Sonnet~5 resolved 12/25 (48\%)
and Claude Opus~5 16/25 (64\%) at low effort, so both satisfied the gate.

The second campaign evaluated \cc{}~2.1.220 at low, high, and xhigh effort,
paired by task across both models (150 cells). Pre-launch manipulation checks
confirmed that the effort setting changed output volume on this path
(low to xhigh: Sonnet~5, 5{,}253 to 35{,}113 output tokens; Opus~5, 9{,}040
to 21{,}719). At low effort, Sonnet~5 averaged 34.8 turns, 34.2 tool calls,
227 seconds, and approximately 4{,}000 thinking tokens per task, compared
with 6.0 turns, 5.1 tool calls, 17.7 seconds, and near-zero thinking on the
deterministic fixtures in Appendix~\ref{app:adaptive}. Structural difficulty
was positively associated with Sonnet~5 compute (Spearman
$\rho\approx0.45$--$0.49$ against gold-patch file count), while Opus~5 showed
little such association.

All 150 cells completed without infrastructure errors. A preregistered budget
cap was amended from \$250 to \$300 before success outcomes were evaluated;
the rationale is recorded in the public audit trail. Increasing effort
consistently increased compute: xhigh versus low added 12--16k thinking
tokens and 23--30 turns and increased billed cost by 2.4--4.5\xd{}. All six
timeouts and all four empty patches occurred at high or xhigh effort.

For Opus~5, resolved counts were 17/25 at low, 18/25 at high, and 16/25 at
xhigh (McNemar xhigh vs.\ low, $p=1.0$), with cost per resolved task of
\$0.75, \$1.89, and \$3.62, respectively. No discordant pair reproduced on
the preregistered second repetition. The two xhigh regressions were
consistent with compute-induced failure modes: one scope-expanding edit
broke previously passing tests, and one run timed out with an empty diff.
Thus, low effort lies on the observed cost--success frontier for Opus~5.

For Sonnet~5, resolved counts were 12/25 at low, 19/25 at high, and 17/25 at
xhigh. The preregistered primary xhigh-versus-low comparison contained six
rescues and one regression ($p=0.125$ after the primary McNemar test; not
significant after Holm correction); the secondary high-versus-low comparison
contained seven rescues and no regressions ($p=0.016$). Cost per resolved
task increased from \$1.76 to \$2.32 to \$3.00, so xhigh is dominated by
high in this sample. On second repetitions, three of six xhigh rescues
reproduced; all three were classified as cross-module-completeness failures
of the low-effort run. Six of ten discordant tasks were resolved by both
effort settings on rerun, indicating substantial run-to-run variability.

A preregistered evidence-triggered escalation rule, simulated on the static
arms, activated on 64--68\% of tasks and did not improve the observed
cost--success frontier: for Sonnet~5 it approximately matched always-xhigh,
and for Opus~5 it was worse than always-low. The campaign therefore shows
that difficult repository-scale tasks can induce substantial computation at
low effort, while additional explicit effort has model-dependent and often
unfavorable cost--success trade-offs. Full protocol and audit materials are
available in the public repository.

\section{Post-Registration DeepSeek Harness Extension}
\label{app:dsh}
This extension was frozen on 2026-08-19 after a baseline activation gate and
before any comparative arm. It is a harness extension of the frozen activation
benchmark: tasks, prompts, fixtures, hidden evaluators, metrics, activation
definition, and statistical procedures are unchanged. The variable is the
agent harness.

\paragraph{Setup.} The harness is DeepSeek Harness (dsh)~0.1.0rc7 using the
bundled unattended \texttt{dsh-jsonrpc-agent} composition. The model is
Claude Sonnet~5, held fixed relative to the \cc{} activation benchmark, served
through a local OpenAI-compatible chain to the first-party Anthropic API. A
body shim removes dsh's budget-less \texttt{thinking} field so LiteLLM can map
\texttt{reasoning\_effort} to the Anthropic thinking budget and clamps
\texttt{max\_tokens} to the model output ceiling; otherwise requests are
captured unchanged. Wire inspection before freezing showed that this evaluated
composition enabled aggressive reasoning on every normal request. The
baseline preserves that shipped behavior.

The dsh runtime is structurally different from \cc{} and \pidev{}. Its
published composition includes an agent spine, local bash and file tools,
subagent delegation, todo state, JSONL persistence/checkpoints, token metering,
and automatic context compaction \citep{deepseekharness}. The broader project
also exposes Code Mode and workflow orchestration, but those capabilities are
not treatment arms in this extension.

\paragraph{Arms and design.} Five arms are evaluated: \texttt{baseline};
\texttt{policy}, which appends the efficiency policy to the dsh system persona;
\texttt{rewrite}, which applies wrapper prompt correction;
\texttt{effort}, which adaptively overrides dsh \texttt{reasoningEffort}
(low/medium/high, with xhigh mapped to high and abstention preserving the
shipped default); and \texttt{full}, combining policy, rewrite, and effort.
There is no runtime-hook arm in this composition. The matrix is 5 tasks
$\times$ 5 arms $\times$ 3 repetitions = 75 valid runs, paired by task and
repetition, with a \$20 cap, 600s timeout, and one retry only for transport
\texttt{no\_json} failures. Tool metrics are parsed from dsh session JSONL
with the same test-pattern and reset-on-edit semantics as the original
activation benchmark.

\paragraph{Activation gate.} Before comparative arms, every induced-waste
task had to exhibit its target behavior under the dsh baseline. The four gate
runs all solved the task and are excluded from comparative analysis:

\begin{table}[ht]\centering\small
\begin{tabular}{lll}
\toprule
Task & Baseline activation evidence & Billed cost \\
\midrule
\texttt{act-testloop-01} & repeated tests without change = 3 & \$0.167 \\
\texttt{act-verify-01} & repeated tests without change = 2 & \$0.340 \\
\texttt{act-approaches-01} & 28,321 output tokens & \$0.639 \\
\texttt{act-explore-01} & repeated tests without change = 1 & \$0.174 \\
\bottomrule
\end{tabular}
\caption{DeepSeek Harness activation gate. On \texttt{act-approaches-01},
the matched \cc{} baseline was approximately 1.25k output tokens and \$0.14;
this cross-harness contrast is diagnostic of activation, not a primary cost
estimand.}
\label{tab:dsh-gate}
\end{table}

A first gate attempt encountered an environment-propagation bug before any
model request was sent. Those four infrastructure-error records were removed,
the runner received a one-line fix, and the gate was rerun before any
comparative arm. This exclusion was disclosed in the frozen extension plan.

\paragraph{Comparative results.} All 75 comparative runs were valid, and all
arms solved all 15 task--repetition cells. Mean billed cost per run was
\$0.2753 for baseline, \$0.2114 for policy, \$0.2109 for rewrite, \$0.0696 for
effort, and \$0.0490 for full. These correspond to point reductions of 23.2\%,
23.4\%, 74.7\%, and 82.2\% relative to dsh baseline. The effort-only arm
captures about 90.9\% of the observed baseline-to-full dollar reduction.
Interventions overlap, so these component differences are not additive. The
rewrite-only point estimate is not promoted as an independently conclusive
inferential result.

\paragraph{Cross-harness interpretation.} The original \cc{} activation
benchmark and the dsh extension use the same closed model and controller logic,
but the dominant lever changes. Under \cc{}, the effort-only arm reduced cost
by about 19\% and the full stack by about 27\%; under dsh the corresponding
figures are about 75\% and 82\%. The contrast identifies a harness--effort
interaction rather than a universal 82\% controller effect. In the evaluated dsh
composition, effort control primarily corrects a high-compute default. The
result also explains why token and cache savings should not be treated as
portable properties of an intervention: dsh and \cc{} expose different
prefixes, tools, context-management behavior, and cache traffic.

\paragraph{Scope.} This extension does not evaluate DeepSeek models, does not
include a dsh runtime-hook arm, and does not establish the effort--success
frontier on hard tasks because the activation fixtures remain ceiling-success
tasks. The direct next experiment is a crossed replication of the frozen
25-instance SWE-bench Verified hard slice under dsh with Sonnet~5 and Opus~5,
using identical low/high/xhigh effort settings and the official Docker judge.
That design can estimate harness$\times$effort, harness$\times$model, and
harness$\times$task-difficulty interactions without changing the task set.

\section{Multiple-Approaches Benefit Audit}
\label{app:ma-audit}
A robustness audit evaluated whether forced branching improves success
(878 branching runs incl.\ paraphrase forms, 821 baselines, 225 paired
blocks; $\sim$14 subgroups, Bonferroni threshold $p<0.004$). Aggregate:
96.5\% vs.\ 95.2\%; discordant blocks 13 favoring branching vs.\ 26
favoring baseline (sign test $p=0.053$, direction negative). The apparent
advantage on non-ceiling blocks (13/1/6) is regression to the mean: its
mirror is larger (26/3/6). On easy cells (pooled baseline $\geq$90\%),
branching reduces success 98.8\% $\to$ 96.4\% ($p\approx.003$, the only
comparison surviving multiplicity, and it is negative) with 2.5\xd{} the
out-of-scope-edit rate. On hard cells (independent definition: pooled
task$\times$model baseline $<$90\%, 17 cells), branching shows
80.5\% $\to$ 85.3\% ($+4.8$pp, Fisher $p=0.427$; $\approx$5 extra
successes), a directional signal only; if taken at face value, each
additional success costs $\approx$7.7\xd{} a full baseline run. Every rescue case was inspected. Several failing baseline runs were very
short non-engagements
(18--40 reasoning tokens), and no discarded branch ever surfaced a
solution that the baseline searched for and missed
(alternatives-implemented is exactly one in every annotated branching
run). Power: at a 96.5\% ceiling, only success effects
$\gtrsim$2.8pp are detectable. By construction the task set contains
essentially no architecture-dependent tasks, so a plausible setting in which branching could help, such as a subtly wrong first obvious fix or two
implementations diverging on hidden behavior, was never tested; a targeted preregistered study with such tasks is needed to evaluate that
regime.

\paragraph{Validation of the hard-cell signal.} The exploratory $+4.8$pp hard-cell effect did not survive robustness
analysis. (i)~\emph{Cross-fitting}: cells
classified hard on screening-phase baselines have held-out baseline
success of 98.3\%; the difficulty label does not reproduce, and the
held-out branching effect is $+0.4$pp (77/78 vs.\ 57/58). (ii)~The
leave-one-out stability (all 21 LOO deltas positive, $+3.5$ to $+9.3$pp),
threshold insensitivity ($+6$ to $+10$pp at cutoffs 0.95--0.80), and the
cell-level difficulty gradient (Spearman $-0.54$) are all computed on the
same data that defines difficulty and therefore inherit the
regression-to-the-mean artifact the cross-fit removes. (iii)~Bootstrap CI
over hard cells: $[-1.7, +12.8]$pp. (iv)~Model consistency is
nemotron-heavy (5/5 positive discordant hard cells) with other models
mixed. (v)~Failure composition: 8/22 hard-cell baseline failures were
degenerate ($<$100 reasoning tokens) or timeouts, reflecting random engagement
rather than problem difficulty; branching's own failures include 4 timeouts.
(vi)~Hidden-test structure: only 5 hidden-only failures exist in hard
cells and branching does not reduce them (2 baseline vs.\ 3 branching);
the theoretical case for branching, namely visible-pass/hidden-fail tasks,
is essentially absent from the benchmark. Accordingly, the hard-cell observation is reported as exploratory and not as
evidence of a branching benefit.


\begin{thebibliography}{20}
\bibitem{repo}
PointFive Labs. \emph{prompt-efficiency-benchmark: benchmark code, task fixtures,
prompt variants, raw ledgers, and analysis}.
\url{https://github.com/PointFiveLabs/prompt-efficiency-benchmark}, 2026.

\bibitem{pi}
Earendil Works. \emph{Pi: an AI agent toolkit and coding-agent CLI}.
\url{https://pi.dev}, 2026.

\bibitem{claudecode}
Anthropic. \emph{Claude Code}. \url{https://claude.com/claude-code}, 2026.

\bibitem{deepseekharness}
DeepSeek AI. \emph{DeepSeek Harness: composable agent runtime and SDK}.
\url{https://github.com/deepseek-ai/deepseek-harness}, 2026.

\bibitem{together}
Together AI. \emph{Inference platform and model catalog}.
\url{https://www.together.ai}, 2026.

\bibitem{litellm}
BerriAI. \emph{LiteLLM proxy (v1.93.0)}.
\url{https://github.com/BerriAI/litellm}, 2026.

\bibitem{kimi}
Moonshot AI. \emph{Kimi K3 technical blog}.
\url{https://www.kimi.com/blog/kimi-k3}, 2026.

\bibitem{baherwani2026}
V.~Baherwani, T.~Goldstein, and A.~Panda.
\emph{Not All LLM Reasoning Is Visible in the Chain-of-Thought}.
arXiv:2607.22925, 2026.

\bibitem{sun2026when2tool}
C.-E.~Sun, L.~Liu, G.~Yan, Z.~Wang, and T.-W.~Weng.
\emph{LLM Agents Already Know When to Call Tools: Even Without
Reasoning} (When2Tool). arXiv:2605.09252, 2026.

\bibitem{liu2025costbench}
J.~Liu, C.~Qian, Z.~Su, Q.~Zong, S.~Huang, B.~He, and Y.~R.~Fung.
\emph{CostBench: Evaluating Multi-Turn Cost-Optimal Planning and
Adaptation in Dynamic Environments for LLM Tool-Use Agents}.
arXiv:2511.02734, 2025.

\bibitem{wu2024catp}
D.~Wu, J.~Wang, Y.~Meng, Y.~Zhang, L.~Sun, and Z.~Wang.
\emph{CATP-LLM: Empowering Large Language Models for Cost-Aware Tool
Planning}. arXiv:2411.16313, 2024.

\bibitem{yang2026efficient}
X.~Yang, L.~Li, H.~Zhou, T.~Zhu, X.~Qu, Y.~Fan, Q.~Wei, R.~Ye, L.~Kang,
Y.~Qin, D.~Liu, Q.~Li, N.~Ding, S.~Chen, and J.~Shao.
\emph{Toward Efficient Agents: Memory, Tool Learning, and Planning}.
arXiv:2601.14192, 2026.

\bibitem{zheng2026corvus}
M.~Zheng, D.~O'Brien, S.~Cui, P.~Pashakhanloo, R.~Mukherjee, M.~Kim, and
S.~Kuhar.
\emph{CORVUS: Context Optimization and Reduction Via Underlying
Synchronization for LLM Coding Agents}. arXiv:2607.22711, 2026.
\end{thebibliography}
\end{document}